\documentclass[lettersize,journal]{IEEEtran}
\usepackage{amsmath,amsfonts}
\usepackage{algorithmic}
\usepackage{algorithm}
\usepackage{array}
\usepackage[caption=false,font=normalsize,labelfont=sf,textfont=sf]{subfig}
\usepackage{textcomp}
\usepackage{stfloats}
\usepackage{subfig}
\usepackage{verbatim}
\usepackage{graphicx}
\usepackage{subcaption}
\usepackage{booktabs}
\usepackage{caption}
\usepackage{cite}
\usepackage{tabularx}
\usepackage{url}
\usepackage[T1]{fontenc}

\begin{document}

\title{Socially-Aware Autonomous Driving: Inferring Yielding Intentions for Safer Interactions}

\author{
    Jing Wang, 
    Yan Jin,~\IEEEmembership{Member,~IEEE,}
    Hamid Taghavifar,~\IEEEmembership{Senior Member,~IEEE,}
    Fei Ding,~\IEEEmembership{Senior Member,~IEEE,}
    Chongfeng Wei,~\IEEEmembership{Member,~IEEE,}

\thanks{Jing Wang, Yan Jin are with the School of Mechanical and Aerospace Engineering, Queen's University Belfast, Belfast, United Kingdom (email: jwang61@qub.ac.uk, y.jin@qub.ac.uk)}
\thanks{Hamid Taghavifar is with the Department of Mechanical and Industrial Engineering, Concordia University, Montreal, QC H3G 1M8, Canada (email: hamid.taghavifar@concordia.ca)}
\thanks{Fei Ding is with the State Key Laboratory of Advanced Design and Manufacturing Technology for Vehicle, College of Mechanical and Vehicle Engineering, Hunan University, Changsha 410082, China  (email: dingfei@hnu.edu.cn)}
\thanks{Chongfeng Wei is with James Watt School of Engineering, University of Glasgow, Glasgow, G12 8QQ, United Kingdom (email: chongfeng.wei@glasgow.ac.uk)}}
\markboth{}%
{Shell \MakeLowercase{\textit{et al.}}: A Sample Article Using IEEEtran.cls for IEEE Journals}
\maketitle
\IEEEpubid{}

\begin{abstract}
Since the emergence of autonomous driving technology, it has advanced rapidly over the past decade. It is becoming increasingly likely that autonomous vehicles (AVs) would soon coexist with human-driven vehicles (HVs) on the roads. Currently, safety and reliable decision-making remain significant challenges, particularly when AVs are navigating lane changes and interacting with surrounding HVs. Therefore, precise estimation of the intentions of surrounding HVs can assist AVs in making more reliable and safe lane change decision-making. This involves not only understanding their current behaviors but also predicting their future motions without any direct communication. However, distinguishing between the passing and yielding intentions of surrounding HVs still remains ambiguous. To address the challenge, we propose a social intention estimation algorithm rooted in Directed Acyclic Graph (DAG), coupled with a decision-making framework employing Deep Reinforcement Learning (DRL) algorithms. To evaluate the method's performance, the proposed framework can be tested and applied in a lane-changing scenario within a simulated environment. Furthermore, the experiment results demonstrate how our approach enhances the ability of AVs to navigate lane changes safely and efficiently on roads.

\end{abstract}


\section{Introduction}
\IEEEPARstart{A}{utonomous} driving decision-making is a critical component of autonomous driving systems, aiming to make reasonable and safe driving decisions based on environmental perception \cite{Malik2022HowDA}. The decision-making process not only needs to consider the kinematic and dynamic constraints of the vehicle but also needs to comply with traffic rules, evaluate potential risks, and coexist safely with other traffic participants in complex driving scenarios, such as executing lane changes on highways and navigating intersections, as illustrated in Figure 1. 
Executing lane changes on the highway remains a formidable challenge for AVs in the real world, primarily due to environmental complexity and uncertainty. The uncertainty primarily stems from two key sources: the inherent noise in sensor data and the inability to directly measure the intentions of human drivers\cite{Hubmann2017DecisionMF}. 
\begin{figure}[htbp]
	\centering
	\includegraphics[width=1.0\linewidth]{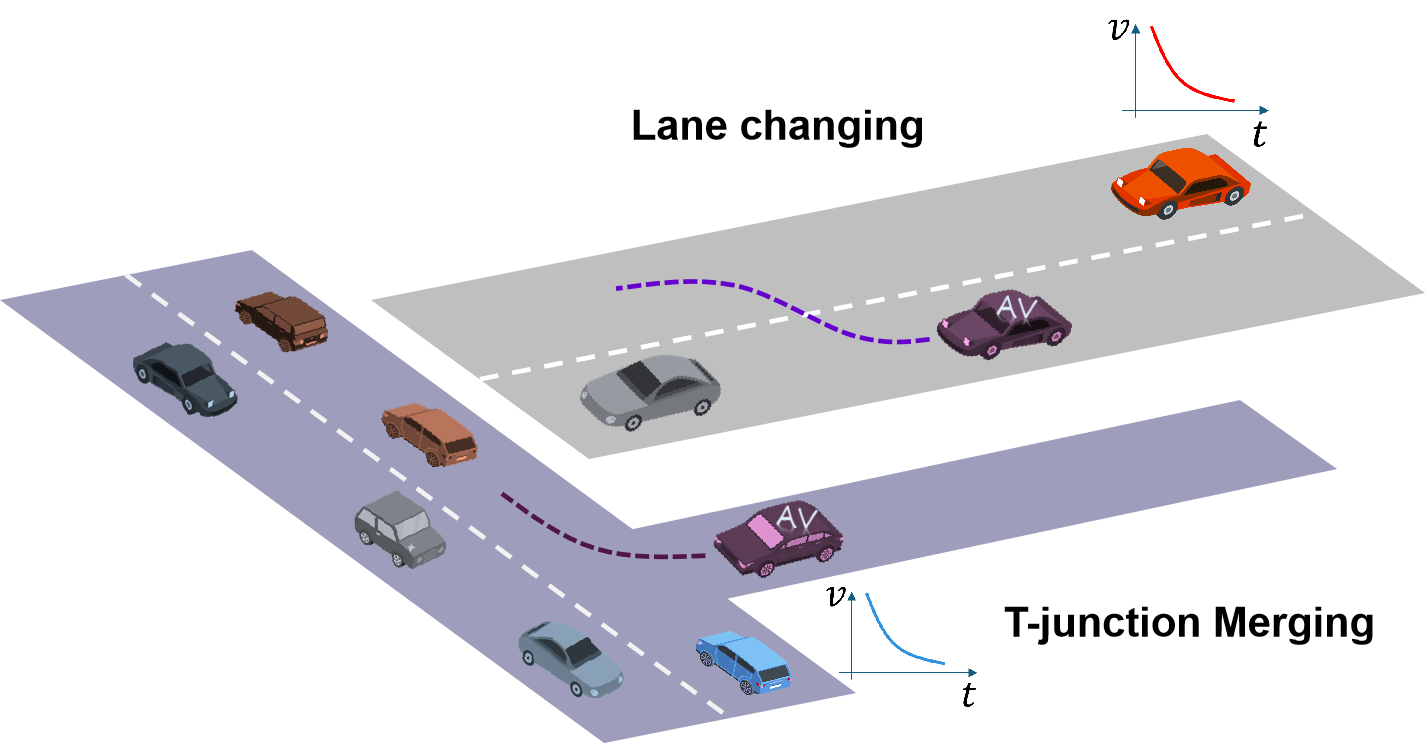}
	\caption{Road Scenarios: Highway Lane Change and T-junction Navigation.}
	\label{fig:01}
\end{figure}
Therefore, mitigating uncertainties is a critical prerequisite for enhancing the reliability of decision-making processes in lane-change scenarios, particularly given the unpredictable behavior of surrounding HVs. Since HVs' behaviors are not straightforward to comprehend based solely on observations during interactions, accurately estimating their intentions is essential for improving the decision-making capabilities of AVs, thereby enhancing overall safety and operational efficiency. The development of efficient intention estimation algorithms is a critical research focus in autonomous driving, as such algorithms not only facilitate the understanding of HVs' current behaviors but also provide insights into their future motion. Moreover, distinct intentions correspond to specific motion patterns, which form the basis for informed and effective decision-making in complex driving scenarios \cite{Cui2023PassingyieldingIE}.
Most studies primarily focus on identifying the merging intentions of HVs \cite{Guo2021DriverLC,Jin2011DriverIR,Li2016LaneCI}. However, in merging scenarios, human drivers have the option to use turn signals to explicitly communicate their intentions to surrounding vehicles, whereas yielding and passing intentions are not as easily conveyed, making their recognition more challenging.\\
In this paper, we address the aforementioned challenge by adopting a Bayesian Network (BN), a DAG model that integrates Bayesian inference with graph-based representations. Additionally, the Bayesian Network integrates the concept of Social Value Orientation (SVO) from social psychology. As a stable personality trait, SVO characterizes how individuals assess and distribute resources between themselves and others in interdependent scenarios, thereby facilitating a more nuanced representation of social interactions in complex driving environments \cite{Hu2017SocialVO}.
The growing complexity of decision-making in autonomous driving has led to a surge in studies leveraging DRL to provide effective solutions \cite{Ma2020ReinforcementLF,Wu2021DeepRL}. This algorithm represents a specialized area of machine learning that synergistically combines reinforcement learning strategies with the representational power of deep neural networks. These networks are instrumental in estimating value or policy functions, thereby capturing the intricate dependencies between environmental states, agent actions, and corresponding rewards. This capability allows agents to derive optimal strategies within high-dimensional and complex state spaces. Consequently, this study adopts the Deep Q-Network (DQN) as the DRL algorithm for decision-making. \\
Specifically, by integrating the concept of SVO into the BN, this study develops a social intention estimation model capable of accurately inferring the probability of yielding intentions among surrounding HVs. This model is seamlessly integrated with the DQN algorithm to systematically model the decision-making process for autonomous driving lane changes. Through this combined approach, AVs can effectively interpret the yielding and passing intentions of surrounding HVs, thereby achieving an optimal balance among safety, efficiency, and smoothness in their lane-changing maneuvers.\\
The specific contributions of this study can be summarized as follows:\\
1. The concept of SVO is embedded within a DAG to model and infer the passing and yielding intentions of surrounding HVs during lane-change conflict situations.\\
2. A novel decision-making framework is proposed, which integrates a probabilistic social intention estimation module. This module estimates the driving intentions of neighboring traffic participants, and the resulting intention probabilities are utilized as part of the state representation in a DRL policy for autonomous lane-change planning.\\
3. The proposed method is implemented and evaluated in a simulated highway driving environment. The results demonstrate that the framework outperforms existing baselines in terms of safety and efficiency during lane-changing maneuvers.\\
The reminder of this paper is organized as follows: Section II provides a comprehensive review of related work on decision-making algorithms, driving intention estimation methodologies, and social psychological models. 
Section III defines lane change scenarios and  introduces a systematic approach to vehicle identification.
Section IV introduces the methods employed in this study.
Section V discusses the selected real-world dataset and outlines the data processing and labeling procedures.
Section VI demonstrates the establishment of experimental scenarios.
Section VII presents an in-depth analysis of experimental results.
Section VIII presents the conclusion.
\section{Related Work}
This section is organized as follows. Firstly, it provides a comprehensive review of state-of-the-art decision-making techniques utilized in lane-change scenarios for AVs. Secondly, it examines intention inference models within the domain of autonomous driving. Finally, it explores relevant social psychological models that contribute to a deeper understanding of decision-making in autonomous driving contexts.
\subsection{Lane Change Decision-Making Models}
Decision-making methods for AVs can be broadly categorized into classical methods and learning-based methods \cite{Liu2021DecisionMakingTF}. Classical methods, such as rule-based approaches have limitations in handling the complexity and uncertainty of real-world driving scenarios. On the other hand, learning-based methods have shown promising performance in decision-making for AVs, and they can learn from data and adapt to dynamic driving environments, making more robust and intelligent decisions. \\
DRL algorithms have emerged as a promising learning-based approach for developing decision-making systems capable of navigating complex environments. Several studies have used DRL algorithms for high-level lane change decision making. DRL was extensively trained in lane change scenarios, and the trained agent autonomously learns and generates an intelligent decision-making function\cite{Hoel2018AutomatedSA}. Ghimire et al. \cite{Ghimire2021LaneCD} used a rule-based DQN where the agent first decides to change lanes using DQN, and then follows rules to execute the maneuver safely. Wu et al. \cite{Wu2022LaneCD} proposed a DRL architecture that incorporates driver inputs as part of the state representation. These applications underscore the versatility and effectiveness of DRL in improving autonomous vehicle technologies.
However, the applications of DRL to lane change decision-making face limitations when interacting with HVs, as DRL requires extensive trial-and-error exploration of the environment to collect experience data. Understanding the potential intent information of surrounding HVs allows the DRL agent to more effectively gather valuable experiences, thereby enhancing sample efficiency and improving the learning process.
\subsection{Driving Intention Estimation}
For AVs, inferring the intentions of other traffic participants like HVs is crucial for safe decision-making, as it enables AVs to better understand and anticipate their' goals from their actions. Previous research efforts have concentrated on two distinct approaches: inferring intents implicitly and explicitly communicating intents. Inferred intents typically pertain to the goals or plans of vehicles, recognized based on their historical activity patterns. In contrast, shared intent is future-oriented, revealing a vehicle's forthcoming intentions before they are manifested through its overt actions \cite{Mahajan2023IntentAwareAD}. The proactive communication of intent between autonomous vehicles and other traffic participants is heavily reliant on vehicle-to-vehicle (V2V) \cite{Demba2018VehicletoVehicleCT} and vehicle-to-infrastructure (V2I) \cite{Butakov2016PersonalizedDA} communication technologies. However, these technologies currently lack maturity, posing a significant challenge for the widespread and effective implementation of intent sharing capabilities.\\
Therefore, we focus solely on inferring the intentions of other surrounding HVs rather than obtaining their sharing intentions. Current research explored various techniques for intention inference such as logic-based approaches, probabilistic methods, and machine learning techniques. Probabilistic and machine learning models are common to be utilized to infer intents of surrounding HVs from their observable behaviors and motion patterns. For example, in \cite{Dong2017IntentionEF,Dong2017InteractiveRM,Dong2018SmoothBE}, a Probabilistic Graph Model was utilized to estimate the merging intentions of surrounding vehicles based on their observed behaviors. A Dynamic Bayesian Network model was proposed to estimate the passing and yielding intentions of the surrounding vehicles by combining the semantic representations with observed vehicle behaviors and features like relative distance, speed, acceleration, and time-to-collision \cite{Cui2023PassingyieldingIE}. In \cite{Song2016IntentionAwareAD}, A decision-making framework has been developed to enable AVs to effectively navigate in complex urban environments by interpreting the intentions of surrounding traffic participants and making informed choices that emphasize safety, efficiency, and socially cooperative behavior. To support this process, a deterministic Hidden Markov Model was introduced, capable of predicting both high-level motion intentions, such as turning left, turning right, or proceeding straight and low-level interactive behaviors, including yielding intentions at intersections. Kherroubi et al. \cite{Kherroubi2022NovelDS} used artificial Neural Network (ANN) to develop yielding intention estimation model based on their current states and vehicle-to-vehicle data. Furthermore, they demonstrated that the estimation model based on ANN achieved an average accuracy and precision above 99 percent, outperforming other baseline methods. A intention-integrated Prediction and Cost function-Based (iPCB) framework utilized an intention prediction module based on Bayesian theorem to infer the yielding intentions of surrounding vehicles \cite{Wei2013AutonomousVS}. Nevertheless, the current models do not take into account the social preferences exhibited by other surrounding HVs, as social behavior on the roads can be attributed to the interplay between altruistic and individualistic characteristics, which jointly shape the behavioral intentions of HVs.
\subsection{Social Preference}
SVO has been utilized as a metric to quantify the social preferences of individual drivers, reflecting their propensity to exhibit either prosocial or egoistic behaviors while operating a vehicle \cite{Schwarting2019SocialBF}. Recently, several efficient, reliable, and valid methods have emerged for measuring SVO, including an online representational approach that infers an individual's preferences from their observed decision-making behaviors \cite{Murphy2011MeasuringSV}. Another widely adopted technique is the SVO Slider Measure \cite{Research1997DevelopmentOP}, comprising six primary items and nine selectable items, or alternatively, a discrete triple dominance measure. However, the most prevalent method to characterize SVO is the ring measure \cite{Liebrand1988TheRM}. Wang et al. \cite{Wang2021ComprehensiveSE} proposed an interaction-aware safety evaluation framework for highly automated vehicles in the roundabout entering scenario, and models the primary other vehicles as game-theoretic agents using level-k game theory and SVO concepts to capture a diverse range of interactive patterns. A optimal control approach for connected and automated vehicles was presented in mixed traffic by incorporating SVO and game theory, aiming to achieve a balance between individual and cooperative driving behaviors \cite{Le2022ACO}. Luca et al. \cite{Crosato2022InteractionAwareDF} integrated a DRL algorithm with the concept of SVO to model the interactive behavior between an autonomous vehicle and a pedestrian. Tong et al. \cite{Tong2023HumanlikeDA} employed SVO as a metric to quantify the social interactions between AVs and HVs, and its integration into a RL framework for human-like decision-making at unsignalized intersections, enabling safer and more socially-compliant autonomous driving in mixed traffic environments.

\section{Problem Statement}
In a lane-change scenario, the autonomous vehicle (AV) in the rightmost lane is constrained by the leading vehicle of the AV and unable to improve its current situation. As a result, the AV initiates a lane-change maneuver and begins to estimate the yielding and passing intentions of surrounding vehicles in the adjacent lane. This estimation leads to two potential decisions for the AV: either proceed with the lane change or remain in the current lane. To provide clarity on the designed lane-change scenario, all vehicles relative to the AV are regarded as "surrounding vehicles," as illustrated in Figure 2. 
\begin{figure}[htbp]
	\centering
	\includegraphics[width=1.01\linewidth]{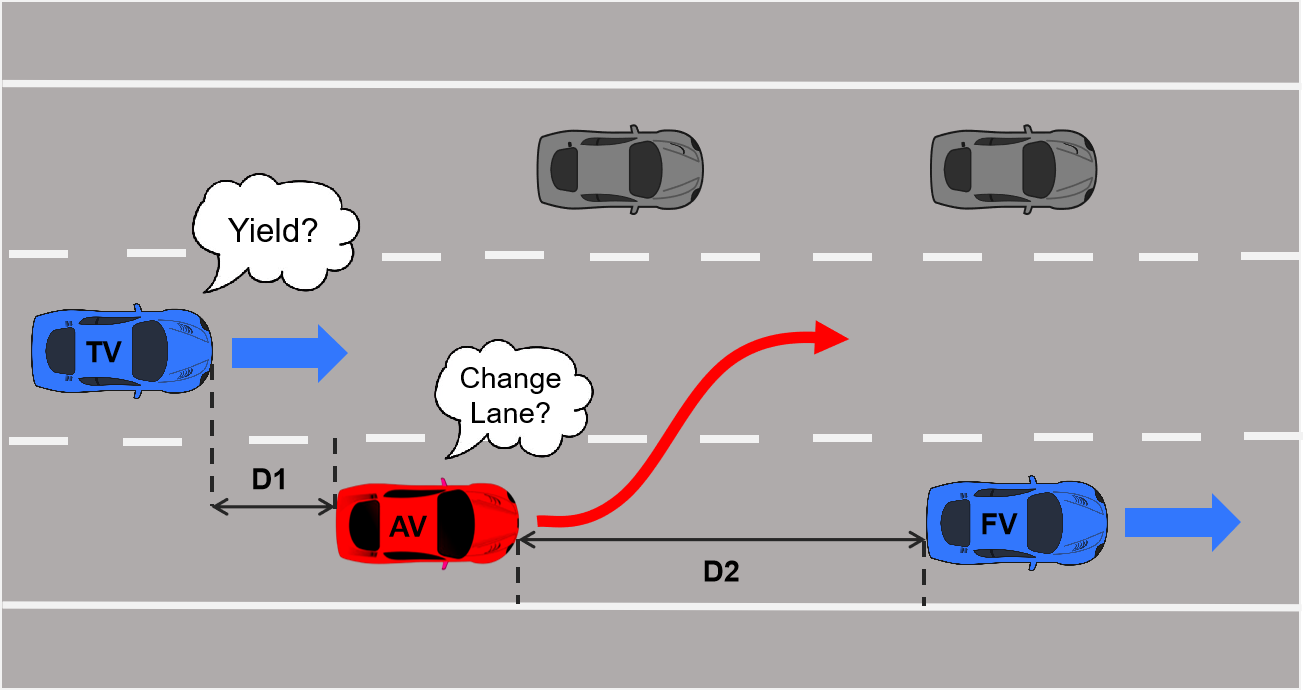}
	\caption{The lane-changing scenario.}
	\label{fig:01}
\end{figure}
Consequently, the analysis focuses primarily on two categories of surrounding vehicles that have a direct influence on the AV decision-making process.
\begin{itemize}
\item[$\bullet$] The target vehicle (TV) is defined as one of the nearest vehicles in the adjacent lane that poses a potential collision risk to the AV during a lane-change maneuver. The driving intention of the TV can be classified into two distinct categories: yielding or passing intention. The TV is positioned to the left and behind the AV before the initiation of the lane-change process, as illustrated in Figure 2.
\end{itemize}
\begin{itemize}
\item[$\bullet$] The forward vehicle (FV) is defined as one of the nearest vehicles positioned ahead of the AV in the target lane. Prior to the initiation of the lane-change maneuver, the leading vehicle is positioned directly in front of the AV, as shown in Figure 2.
\end{itemize}
In this paper, our primary focus is on analyzing the lane-changing behavior of the AV. We examine the entire interaction process, from the initiation of the lane-change maneuver to its successful completion. Specifically, at the onset of the lane-changing process, the AV initiates the maneuver to improve its own traffic situation due to the slow velocity of the FV in the current lane. Following this initial moment, the AV begins to estimate the intentions of the TV in the adjacent lane, enabling it to make a reliable lane-change decision. Once the TV demonstrates a yielding intention, the AV proceeds with the lane change. Finally, at the completion point of the maneuver, the AV successfully transitions into the target lane.

\begin{figure*}[htbp]
	\centering
	\includegraphics[width=1.0\linewidth]{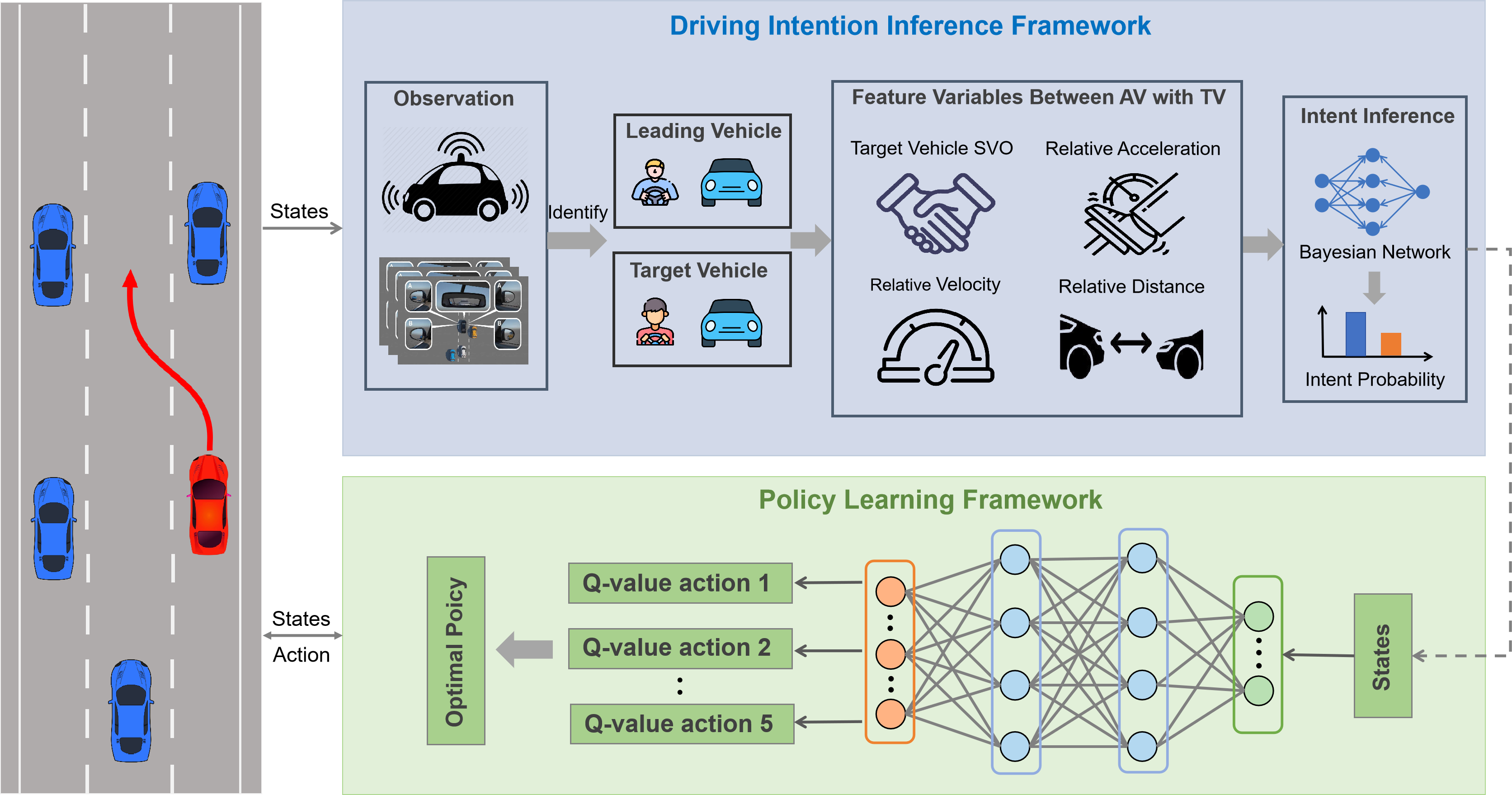}
	\caption{A socially-aware decision-making framework for autonomous lane-change maneuvers.}
	\label{fig:01}
\end{figure*}
\section{Methodology}
In this section, we present a lane-change decision-making framework that takes into account the driving intentions of the TV in the adjacent lane, as illustrated in Figure 3. Firstly, we give a detailed description of the social psychological model SVO. Next, we introduce a social intention estimation model developed by integrating a constructed BN with the SVO framework. Finally, we present a decision-making framework, which explicitly considers the driving intentions of the TV.
\subsection{Social Value Orientation}
Modeling social interactions is significant to the decision-making process of vehicle behavior, as it allows for the assessment of actions that impact both the individual and others. Social value, therefore, becomes an essential metric for evaluating these outcomes, ensuring that decisions made by vehicles not only optimize individual benefits but also take into account the welfare of others within the environment [28]. SVO is employed to characterize social interaction behaviors among vehicles and to extract the SVO of the TV. 
In this context, an individual’s utility function, which combines the utilities of both the self and others, serves as an effective representation of their social preferences.
\begin{equation}
U=\cos(\varphi)U_{ego}+\sin(\varphi)U_{other}
\end{equation}
where \(\varphi\) is the social personality of each individual with the SVO value. \(U_{ego}\) denotes the final utility of the ego agent, while \(U_{other}\) is the final utility of the others. However, in the area of autonomous driving, few studies have focused on extracting the social preferences of surrounding HVs based on real-world dataset. Consequently, the function has been revised and incorporated into our research as follows:\\ 
\begin{equation}
\varphi_i(t)=arctan(\frac{\Delta U_j(t)}{\Delta U_i(t)})
\end{equation}
where \(\varphi_i(t)\) of the above-mentioned equation characterizes the SVO value of the TV, thereby representing varying driving social preferences over time. \(\Delta U_i(t)\) and \(\Delta U_j(t)\) represent the changes in longitudinal position of the TV and its neighboring vehicles between two consecutive time points, respectively. Generally, the arithmetic mean is utilized to calculate the final utility of other surrounding vehicles \cite{McKee2020SocialDA}:\\
\begin{equation}
\Delta U_j(t)=\frac{1}{n-1}\sum_{k\in\eta(j)}\Delta U_k(t)
\end{equation}
where \(\Delta U_k(t)\) represents the final utility of the \(k^{th}\) vehicle among other surrounding vehicles of TV, n is the number of vehicles, and \(\eta(j)\) is the index set of the neighboring vehicles of the TV.
\begin{figure}[htbp]
	\centering
	\includegraphics[width=0.7\linewidth]{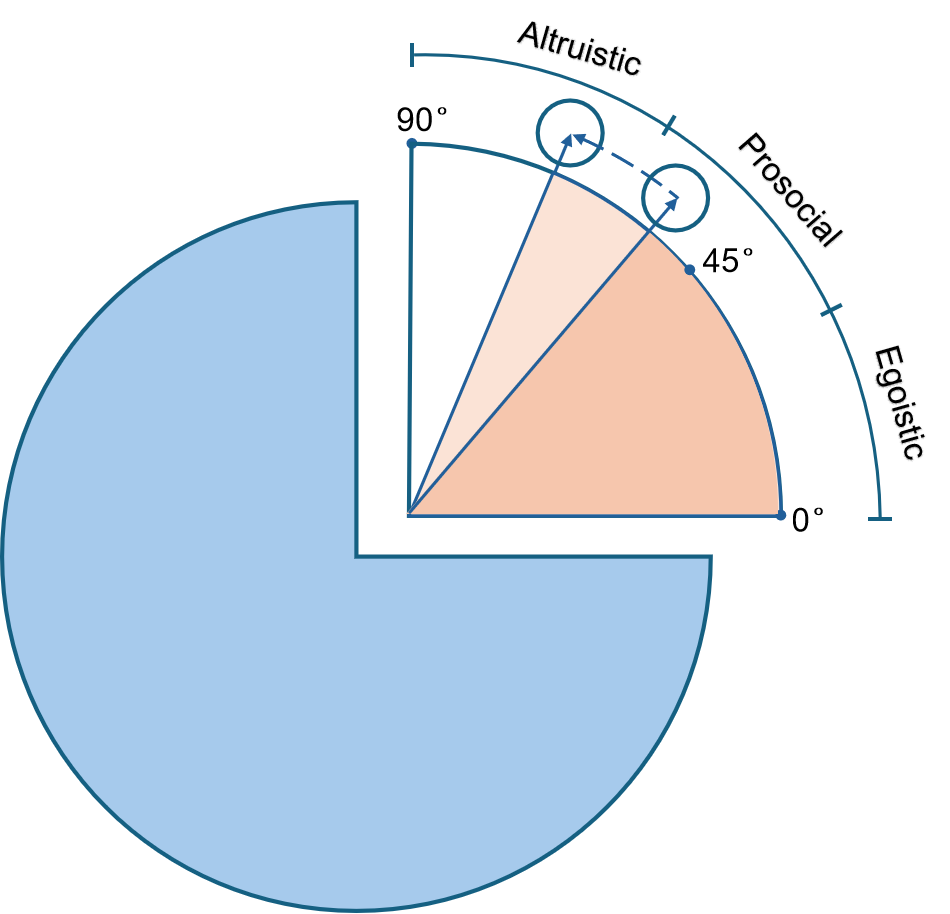}
	\caption{The SVO ring.}
	\label{fig:01}
\end{figure}
In our study, we consider only the AV as the neighbor of the TV. 
Typically, the SVO value is characterized by a ring angle ranging from 0 degrees to 360 degrees. However, to simplify our research, the SVO angle has been limited to a range from 0 degrees to 90 degrees, thereby excluding more complex relevant behavioral properties. This simplified representation is visually illustrated in Figure 4.
\subsection{Social Intention Estimation Model}
\subsubsection{Bayesian Network}
Understanding the behaviors of surrounding HVs, particularly in inferring their driving intentions, is a critical for enabling AVs to determine the optimal timing for lane changes. To achieve this, a BN is employed to estimate the yielding and passing intentions of the TV. 
\begin{figure}[H]
	\centering
	\includegraphics[width=0.65\linewidth]{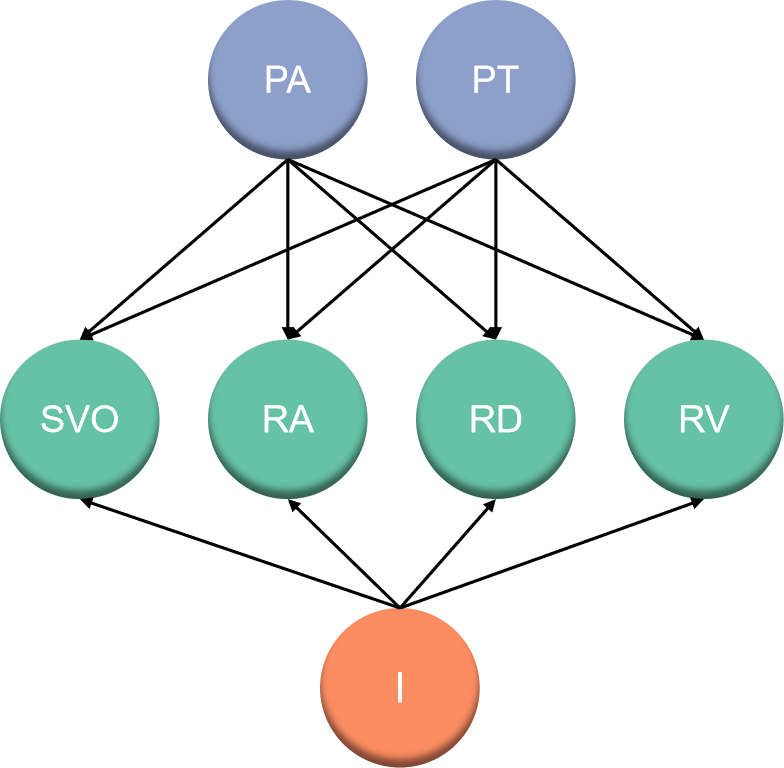}
	\caption{The architecture of Bayesian Network.}
	\label{fig:01}
\end{figure}
A BN is a probabilistic graphical model that captures conditional dependencies among random variables through a DAG, denoted as \(\mathbf{\mathcal{G}} = (\mathbf{\mathcal{V}}, \mathbf{\mathcal{E}})\)
. In this graph, the set of vertices \(\mathbf{\mathcal{V}} = X = \{X_{1},X_{2},....,X_{i}\}\) corresponds to random variables, and both variable nodes are connected by a directed edge \(\mathbf{\mathcal{E}}\) indicate that the connected random variable nodes have a causal relationship. For each node \( X_i \) in a Bayesian network, its conditional probability \( P(X_i \mid X_{\text{pa}(i)}) \) depends only on its set of parent nodes \( X_{\text{pa}(i)} \). The joint probability \( P(X) \) can be computed by taking the product of the conditional probabilities of each node.
\begin{equation}
P(X) = \prod p(X_{i}|X_{pa(i)})
\end{equation}
\subsubsection{Structural Design}
The proposed BN structure consists of three hierarchical layers. Additionally, the model incorporates SVO, relative acceleration (RA), relative velocity (RV), and relative distance (RD) between the AV and the TV, as well as the driving intentions (I) of TV, as illustrated in Figure 5.\\
The top layer encapsulates several state variables of both the AV and the TV:
\begin{itemize}
\item[$\bullet$] Feature vector PA: This vector represents the state variables of the AV, including its longitudinal position, velocity, and acceleration.
\end{itemize}
\begin{itemize}
\item[$\bullet$] Feature vector PT: Analogous to PA, this vector encapsulates the state variables of the TV, including its longitudinal position, velocity, and acceleration.
\end{itemize}
The intermediate layer comprises a set of continuous variables, SVO, RA, RV, RD. These variables are derived from the dynamic states between the AV and the TV:
\begin{itemize}
\item[$\bullet$] The variable SVO: The variable is to assess the TV’s social preferences and behavioral tendencies regarding the allocation of resources or outcomes among itself and others. 
\end{itemize}
\begin{itemize}
\item[$\bullet$] The variable RA: The variable represents the relative acceleration between the AV and the TV.
\end{itemize}
\begin{itemize}
\item[$\bullet$] The variable RV: The variable denotes the relative velocity between the AV and the TV.
\end{itemize}
\begin{itemize}
\item[$\bullet$] The variable RD: The variable corresponds to the relative distance between the AV and the TV.
\end{itemize}
The bottom layer contains the driving intentions of TV:
\begin{itemize}
\item[$\bullet$] The variable I: A discrete latent variable represents the TV’s driving intentions, which can assume one of two categorical states: yielding and passing.
\end{itemize}

\subsubsection{Bayesian Inference for Intentions}
The objective of this section is to estimate the likelihood of each possible driving intention. To simplify the representation, the feature vector \(\text{Y} = \{\text{SVO}, \text{RA}, \text{RV}, \text{RD}\}\) is introduced to collectively denote the four continuous variables. For each continuous node in \(\text{Y}\) is modeled using a Gaussian distribution, conditioned on the binary intention variable \(\text{I}\), where (\( I = 1 \)) represents a yielding intention and (\( I = 0 \)) is a passing intention.
\begin{equation}
\begin{aligned}
    P(Y | I = 0) \sim \mathcal{N}(\mu_0, \sigma_0^2)\\
    P(Y | I = 1) \sim \mathcal{N}(\mu_1, \sigma_1^2)
\end{aligned}
\end{equation}
To enable inference through discrete Bayesian methods, the domain of each continuous variable in \( Y \) is partitioned into \( k \) non-overlapping intervals:
\begin{equation}
    Y = \{B_i\}_{i=1}^k, \quad B_i = [l_i, u_i]
\end{equation}
where $l_i$ and $u_i$ represent the lower and upper bounds of the interval respectively, satisfying $l_{i+1} = u_i$.\\
These bounds are derived from the percent-point function (PPF) of the Gaussian distribution, which maps quantiles to their corresponding values:
\begin{equation}
    l_0 = \inf \operatorname{PPF}(p_1), \quad u_k = \sup \operatorname{PPF}(p_2)
\end{equation}
where \(p_1\) and \(p_2\) define the lower and upper percentiles used to truncate the distribution’s support, ensuring coverage of the majority of probability mass. The Percent-Point Function is computed as:
\begin{equation}
    \text{PPF}(p) = \mu + \sigma \cdot \Phi^{-1}(p)
\end{equation}
The probability mass associated with each interval $B_i$ is computed as:
\begin{equation}
P(Y \in B_i) = \int_{l_i}^{u_i} p(y)\,dy \approx F(u_i) - F(l_i)
\end{equation}
Finally, the discrete approximation \(\hat{p}_i\) of the continuous distribution over intervals is obtained by normalization:
\begin{equation}
    \hat{p}_i = \frac{P(B_i)}{\sum_{j=1}^k P(B_j)}
\end{equation}
This process enables the transformation of continuous variables into a discretized form suitable for probabilistic inference in BN.
Given evidence \( E = \{ra = ra_1, rd = rd_1, rv = rv_1, svo = svo_1\} \), the corresponding discrete intervals \(B_e\) are first determined by mapping each continuous variable to its respective discretized bin based on the previously established interval boundaries. These discretized observations are then utilized to perform probabilistic inference within a Bayesian framework. \\
The joint probability of the driving intentions \(I\) and the evidence \(E\) is computed as the product of the conditional probabilities of each continuous variable given the intention \(I\), along with the prior probability of \(I\):
\begin{flalign}
& P(E, I) = P(ra=ra1 \mid I)P(rd=rd1 \mid I) \notag\\
& \quad \quad \quad \quad \times P(rv=rv1 \mid I) P(svo=svo1 \mid I) P(I)
\end{flalign}
Subsequently, the posterior distribution over \(I\) is derived via Bayes’ theorem, which normalizes the joint probability using the marginal likelihood of the evidence \(P(E)\).
\begin{equation}
P(I \mid E) = \frac{P(E, I)}{P(E)} = \frac{P(E \mid I) P(I)}{P(E)}
\end{equation}
Here, \( P(E \mid I) \) is the likelihood of the evidence given the driving intention \( I \), and \( P(E) \) is the marginal likelihood of the evidence, which acts as a normalizing constant to ensure that the sum of posterior probabilities over all possible values of \( I \) equals one.
\begin{equation}
P(E) = \sum P(E \mid I) \cdot P(I)
\end{equation}
The posterior probability of \( I \), given the evidence \( E \), can be further decomposed for each state of the TV's intention, yielding the following expressions for each outcome of \( I \):
\begin{equation}
    \begin{aligned}
        P(I = 1 \mid E) &= \frac{P(E \mid I = 1) \cdot P(I = 1)}{P(E)}\\
        P(I = 0 \mid E) &= \frac{P(E \mid I = 0) \cdot P(I = 0)}{P(E)}
    \end{aligned}
\end{equation}
These two equations represent the computation of the posterior probability of the TV’s intention to yield (\( I = 1 \)) or to pass (\( I = 0 \)), given the observed evidence. 
\subsection{Decision-making Algorithm Based on Deep Reinforcement Learning}
The classical Q-learning algorithm is used for problems with small discrete state and action spaces. However, it lacks generalization capability when the observation and action spaces expand, making it impractical for high-dimensional tasks. To address this issue, a DQN algorithm was proposed, which combines the Q-learning algorithm with deep learning techniques, particularly deep neural networks, to handle reinforcement learning problems in high-dimensional state and action spaces. It mainly approximates the state-action values with a deep neural network \(Q(s, a; \theta)\), where \(\theta\) is the parameter of the network.\\
DQN is mainly designed based on temporal difference learning, which aims to approximate the optimal action-value function:
\begin{equation}
    Q(s_t,a_t)\leftarrow Q(s_t,a_t)+\alpha\cdot\left[r_t+\gamma\cdot\max_aQ(s_{t+1},a)-Q(s_t,a_t)\right]
\end{equation}
The loss function of DQN is designed to minimize the temporal difference error make its predicted Q-values approach the true values more closely, ultimately leading to improved performance in reinforcement learning tasks. The true values can be represented as:
\begin{equation}
    y = R+\gamma\max_{a\in\mathcal{A}(S')}Q(S',a;\theta')
\end{equation}
This error quantifies the discrepancy between what the network predicts the Q-values should be for each action in a given state and what they should ideally be according to the Bellman equation. The Bellman equation provides a target for the network to learn from by recursively defining the optimal action-value function in terms of itself.
\begin{equation}
    L=\mathbb{E}\left[\left(y - Q(S,A;\theta)\right)^2\right]
\end{equation}
Gradient descent can efficiently update the parameters of the neural network of DQN in the direction that decreases the loss function. By iteratively adjusting the parameters based on the gradients of the loss function with respect to each parameter, gradient descent drives the network towards a configuration that better approximates the optimal action-value function:
\begin{equation}
    \nabla_\theta L=\mathbb{E}\left[\left(y - Q(S,A;\theta)\right)\nabla_\theta Q(S,A;\theta)\right]
\end{equation}
\subsection{Markov Decision Process Formulation}
The Markov Decision Process (MDP) stands as a pivotal mathematical construct for modeling complex decision-making scenarios. This framework performs well in environments where outcomes are influenced by both stochastic elements and deliberate actions taken by an agent. It is widely employed in the field of autonomous driving to model various sequential decision-making problems. In this study, the sequential decision-making of the AV can be characterized as a MDP which is typically represented as a four-element tuple \(\langle \mathcal{S}, \mathcal{A}, \mathcal{R}, \mathcal{P} \rangle \). When the AV performs an action at time t \(a_{t}\) \(\in\) \(\mathcal{A}\) based on the current state \(s_{t}\) \(\in\) \(\mathcal{S}\), the state can transition to a new state according to transition probabilities \(\mathcal{P}\)\((s_{t+1}|s_{t},a_{t})\). This transition then results in a reward \(r_{t}\) \(\in\) \(\mathcal{R}\) based on the new observation.
\subsubsection{State Space (\(\mathcal{S}\))} In the context of MDP, the decision-making process is modeled as a sequence of states. Each state represents a specific situation in which the AV operates.  Furthermore, in order to distinguish the AV from other vehicles, the first row of the features matrix is designated to represent the features of the AV.
\begin{equation} \label{eqn2}
  \begin{split}
  \mathcal{S} &= [x^{i}, y^{i}, v_{x}^{i}, v_{y}^{i}]
  \end{split}
\end{equation}
Where the state space includes the \(i^{th}\) vehicle's longitudinal and lateral positions along with the corresponding longitudinal and lateral velocities.

\subsubsection{Action Space (\(\mathcal{A}\))}
The AV uses throttle and braking adjustments to control its speed during a lane-change maneuver, ensuring it reaches the desired velocity. Additionally, it is capable of adjusting the steering angle to navigate towards the target lane by making left or right changes. Therefore, in our model, at every time step, the AV has an option.
\begin{equation}
    \mathcal{A}=\{left,right,constant,accelerated,decelerated\}
\end{equation}

\subsubsection{Reward Function (\(\mathcal{R}\))}
Each state-action pair is associated with a numerical reward signal. This reward \(\mathcal{R}\) represents the immediate benefit or cost of taking a specific action in a particular state. In reinforcement learning-based frameworks, the AV seeks to learn a policy that maximizes the expected return over a temporal horizon.
\begin{align}
     \mathcal R & = r_{t}^{or} * (\omega_{t}^{c}*r_{t}^{c} + \omega_{t}^{rl}*r_{t}^{rl} + \omega_{t}^{e}*r_{t}^{e})
\end{align}
Where \(r_{t}^{c}\), \(r_{t}^{rl}\), \(r_{t}^{or}\), and \(r_{t}^{e}\) represent a collision penalty term, an arrival lane reward term, a on-road reward term, and a target velocity penalty term, respectively. The weights \(\omega\) correspond to their respective rewards are 1.
\begin{itemize}
\item The component \(r_{t}^{c}\) is a collision penalty, in order to enhance the safety of the AV and prevent collisions:
\begin{equation}
    r_{t}^{c}=\begin{cases}-1,&\text{if collision}\\0,&\text{otherwise.}\end{cases}
\end{equation}

\item The component \( r_{t}^{rl} \) represents the reward assigned to the AV based on its lane position. It is formulated as follows:  
\begin{equation}
    r_{t}^{rl} = \frac{\text{lane\_index}}{\max(\text{num\_lanes} - 1, 1)}
\end{equation}  
where \( \text{lane\_index} \) denotes the index of the lane currently occupied by the AV, and \( \text{num\_lanes} \) represents the total number of available lanes.

\item The component \(r_{t}^{or}\) is the reward that ensure that the AV remains on the designated road.
\begin{equation}
    r_{t}^{or}=\begin{cases}1.5,&\text{if on road} \\0,&\text{otherwise.}\end{cases}
\end{equation}

\item The component \( r_{t}^{e} \) serves as a penalty when the AV fails to achieve the desired velocity after executing a lane-change maneuver. It is initially formulated as follows:  
\begin{equation}
X_i = \frac{v_{\text{forward}} - v_{\min}}{v_{\max} - v_{\min}}
\end{equation}  
where \( v_{\text{forward}} \) denotes the current velocity of the AV in the forward direction, while \( v_{\min} = 20 \) m/s and \( v_{\max} = 30 \) m/s define the range for velocity normalization.  

To ensure that the penalty remains within a predefined range \([a, b]\), where \( a = 0 \) and \( b = 1 \), the final reward function is expressed as:  
\begin{equation}
r_{t}^{e} =
\begin{cases} 
a, & X_i < a \\ 
X_i, & a \leq X_i \leq b \\ 
b, & X_i > b 
\end{cases}
\end{equation}  
\end{itemize}

\section{Experimental Dataset and Processing}
\subsection{Naturalistic Dataset}
\begin{figure}[H]
	\centering
	\includegraphics[width=0.9\linewidth]{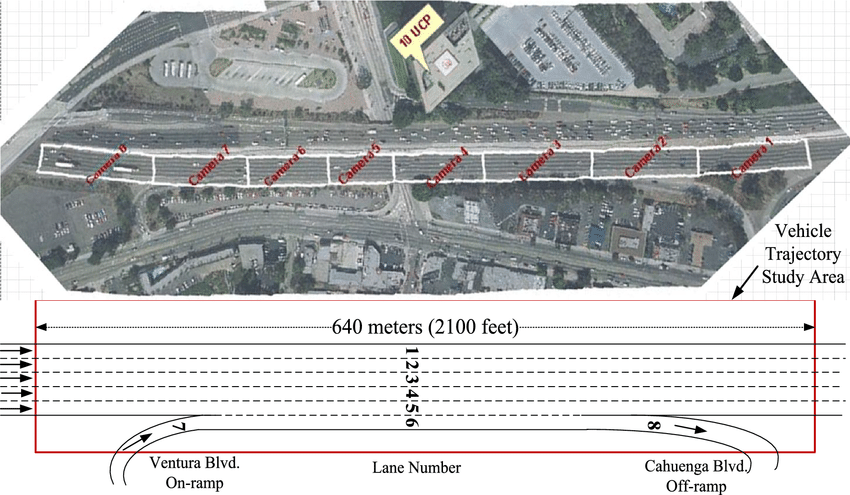}
	\caption{The study area is a segment from the NGSIM U.S. Highway 101 dataset [31].}
	\label{fig:01}
\end{figure}
Currently, the datasets available for studying realistic driving environments encompass a diverse range of scenarios and contexts. These include the Next Generation Simulation (NGSIM) dataset \cite{Alexiadis2004TheNG}, which provides detailed vehicle trajectory data in urban environments; the HDD dataset \cite{Ramanishka2018TowardDS}, renowned for its focus on human driving behaviors; the Argoverse dataset \cite{Chang2019Argoverse3T}, providing richly annotated tracking data for autonomous vehicle research; the highD dataset \cite{Krajewski2018TheHD}, capturing high-definition highway driving scenarios; and the International, Adversarial, and Cooperative moTION (INTERACTION) dataset \cite{Zhan2019INTERACTIONDA}, notable for its inclusion of complex interactive driving situations across various international contexts.\\
Our experiments are conducted using the NGSIM US-101 dataset, a naturalistic driving dataset collected via drone-based video recordings, as depicted in Figure 6. This dataset offers high-resolution vehicle trajectory data, along with comprehensive contextual information, including details about surrounding vehicles involved in lane-change maneuvers. 
\subsection{Data Processing}
There are several steps to process the US-101 dataset:
\begin{enumerate}
\item Filter out all vehicles within lanes from 6 to 8, as well as vehicles classified as types 1 and 3, since our analysis focuses on vehicles traveling in the main lanes from 1 to 5.
\item Identify all vehicles that successfully execute lane changes from their current lane to an adjacent lane, designating them as ego vehicles. 
\item Prior to lane changes, the ego vehicles without leading vehicles in the same lane can also be excluded from consideration.
\item Mark the duration of each lane-changing event from initiation to completion, defining the lane-changing point as the midpoint of the maneuver, with the initiation point set at 30 frames prior and the completion point at 30 frames after, based on the US-101 dataset.
\item Identify all vehicles that interact with those ego vehicles during lane-changing maneuvers and have the potential conflict with the ego vehicles, designating these vehicles as conflicting vehicles.
\end{enumerate}
\subsection{Dataset labeling}
To train the proposed BN model based on pyAgrum \cite{ducamp2020agrum}, the naturalistic US-101 dataset needs to be labeled with passing and yielding intentions of the conflicting vehicles based on two criteria. First, labeling depends on whether the conflicting vehicles yield to ego vehicles, which is determined by whether the ego vehicles complete lane changes first or the conflicting vehicles pass first. Secondly, intentions labeling is based on the comprehensive SVO ring scores calculated throughout vehicle interactions.
\subsubsection{Yielding Scenario}
After data processing and labeling, the final dataset consists of 36 lane-changing conflict cases that meet the specified conditions. Moreover, the dataset includes 16 yielding cases and consists of 976 frames, where each frame represents a sample. The US-101 dataset records vehicle position data at 0.1-second intervals, meaning that each frame corresponds to 0.1 seconds. All samples contain the complete state information of the ego vehicles, leading vehicles, and conflicting vehicles.
\begin{figure}[H]
	\centering
	\includegraphics[width=0.8\linewidth]{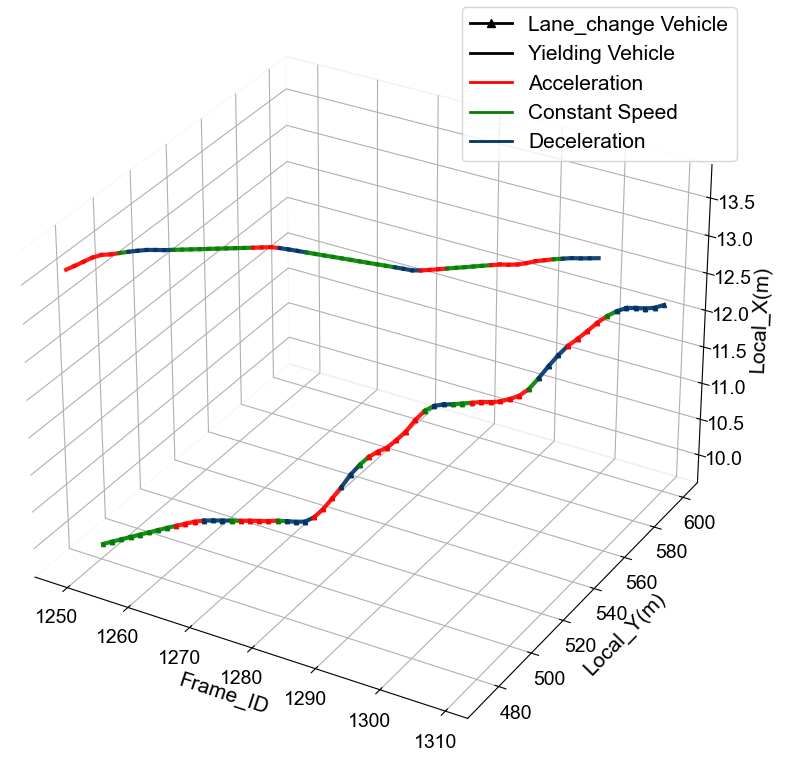}
	\caption{Spatiotemporal Trajectories of Yielding Interaction.}
	\label{fig:01}
\end{figure}
One of yielding cases is illustrated in Figure 7 and Figure 8, where Figure 7 provides a visualization of the lane-change process, highlighting the interaction and the corresponding yielding response of the conflicting vehicle. In this case, the ego vehicle corresponds to vehicle ID 388, and the conflicting vehicle corresponds to vehicle ID 387 in the US-101 dataset. The behaviors of both the ego vehicle and the conflicting vehicle are color-coded to represent three distinct states: accelerated, decelerated, and constant speed. 
\begin{figure}[H] 
    \centering 

    \begin{minipage}{0.24\textwidth}
        \centering
        \includegraphics[width=\textwidth]{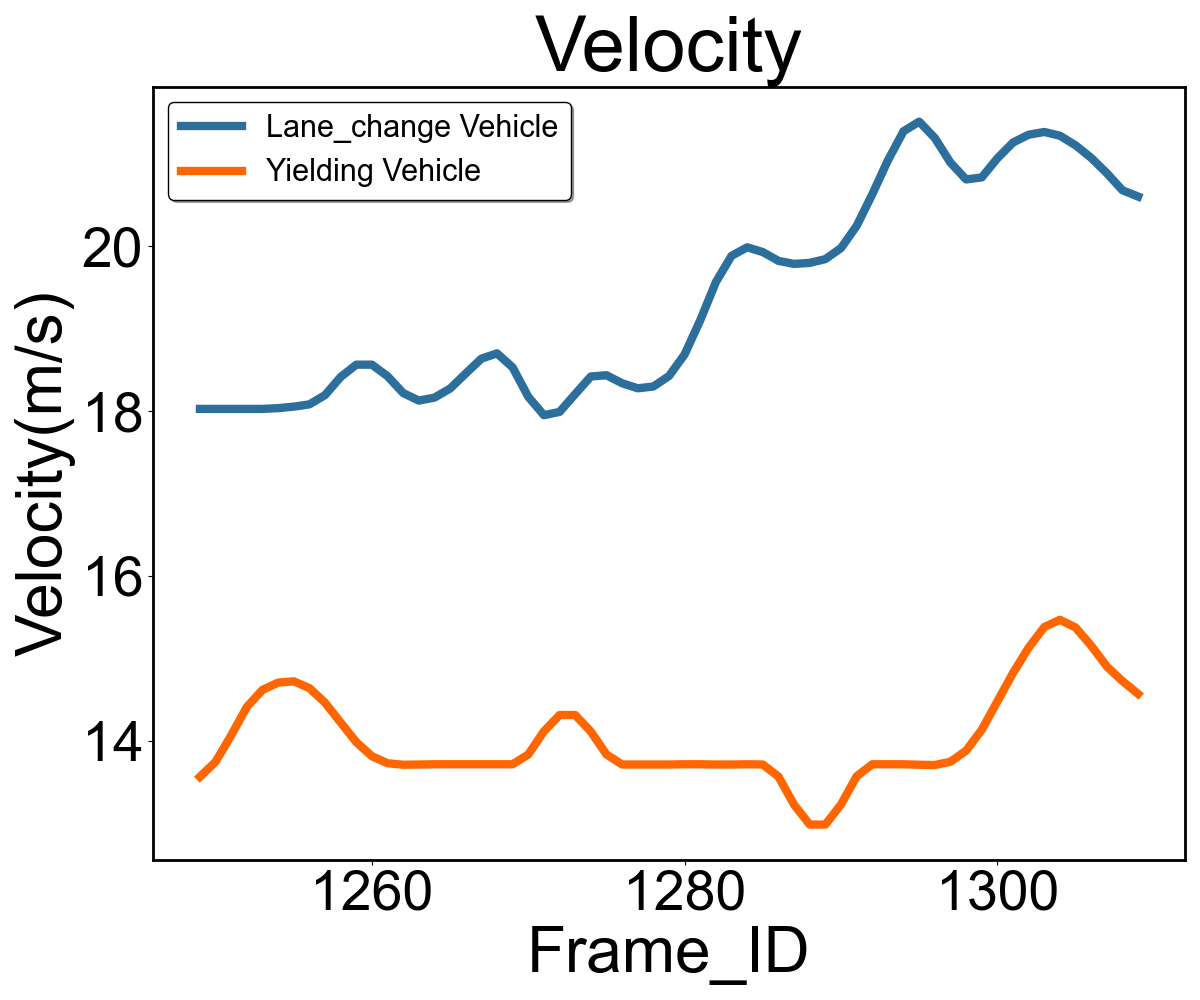} 
    \end{minipage}
    \hfill
    \begin{minipage}{0.24\textwidth}
        \centering
        \includegraphics[width=\textwidth]{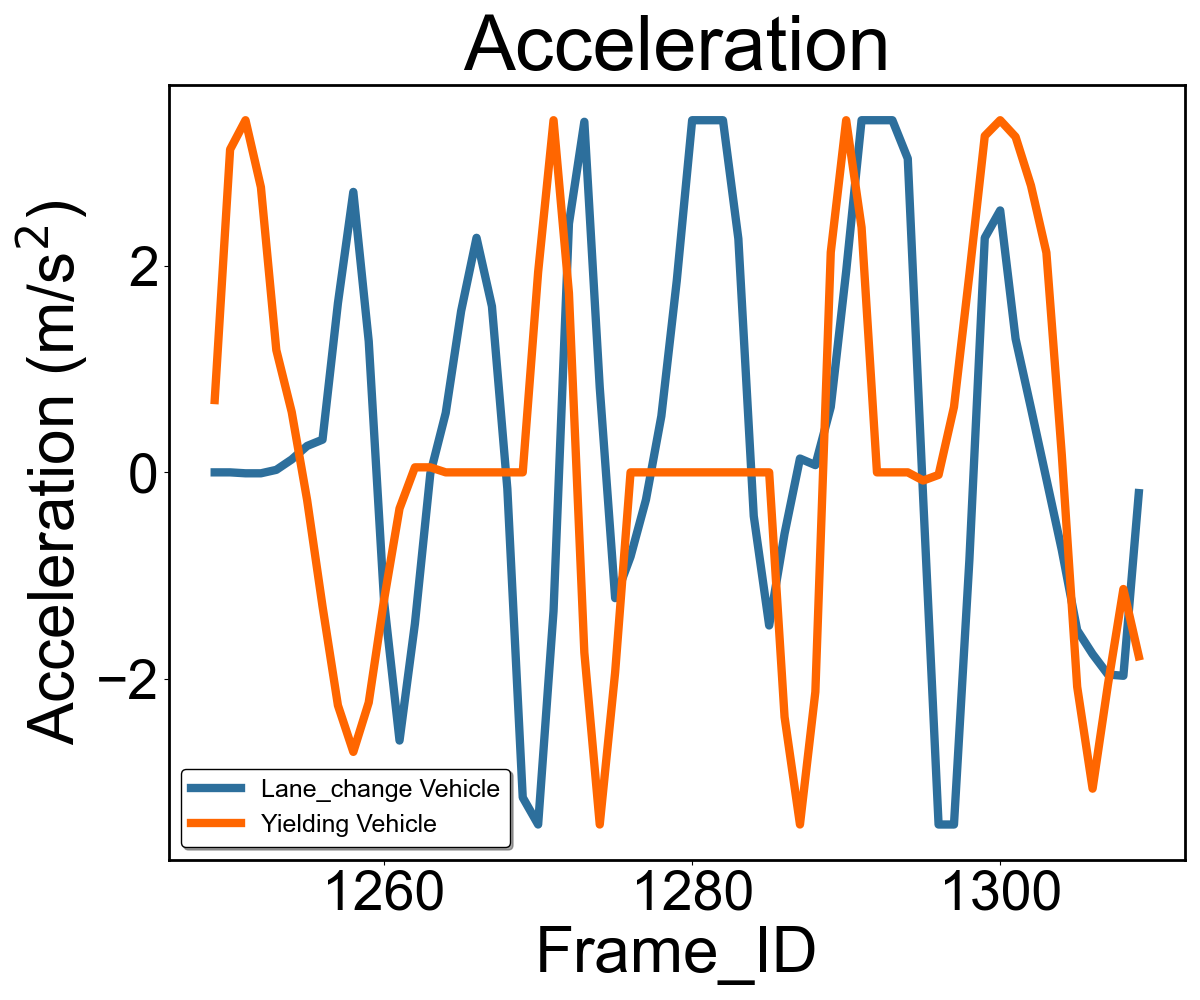} 
    \end{minipage}

    \begin{minipage}{0.24\textwidth}
        \centering
        \includegraphics[width=\textwidth]{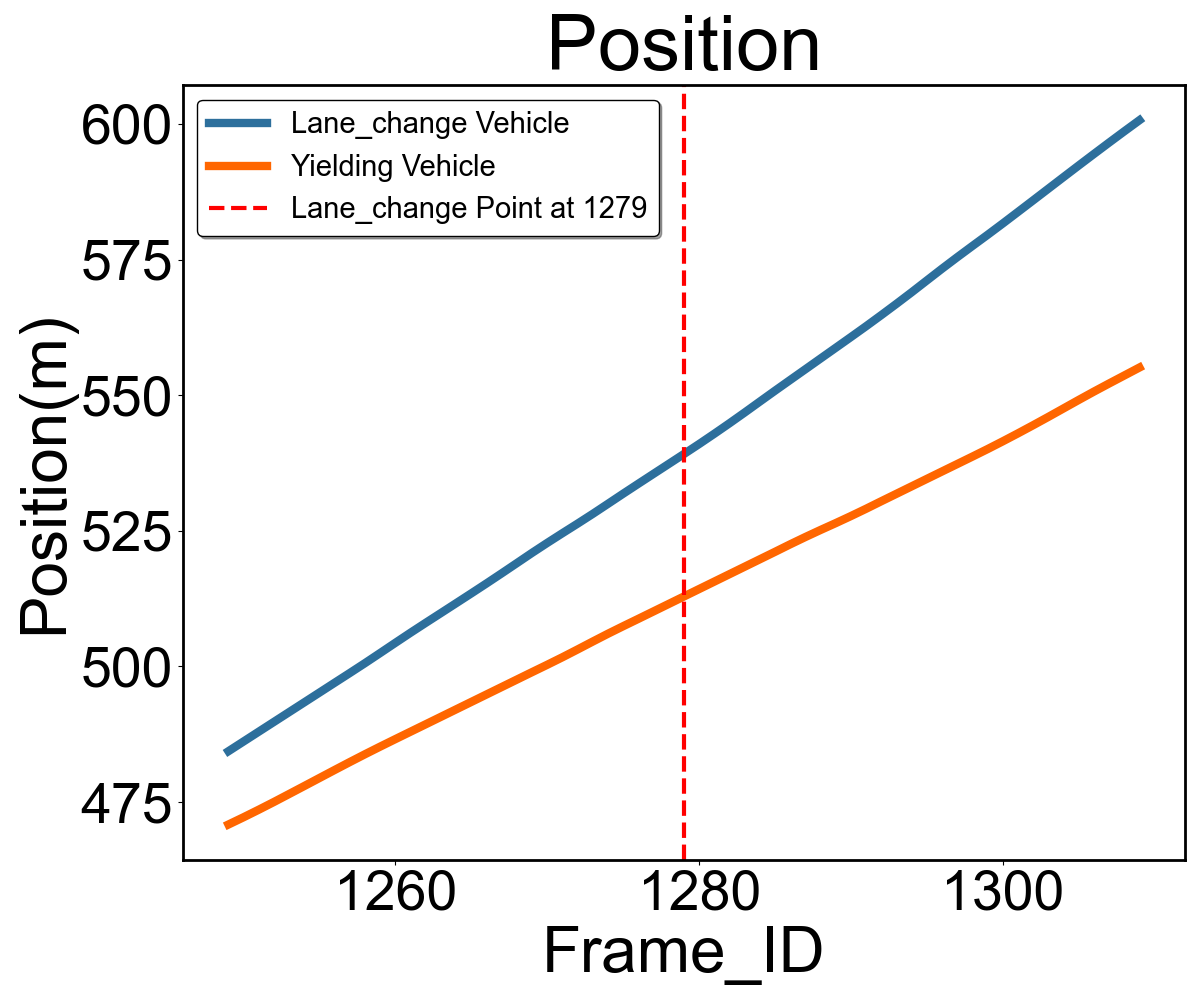} 
    \end{minipage}
    \hfill
    \begin{minipage}{0.24\textwidth}
        \centering
        \includegraphics[width=\textwidth]{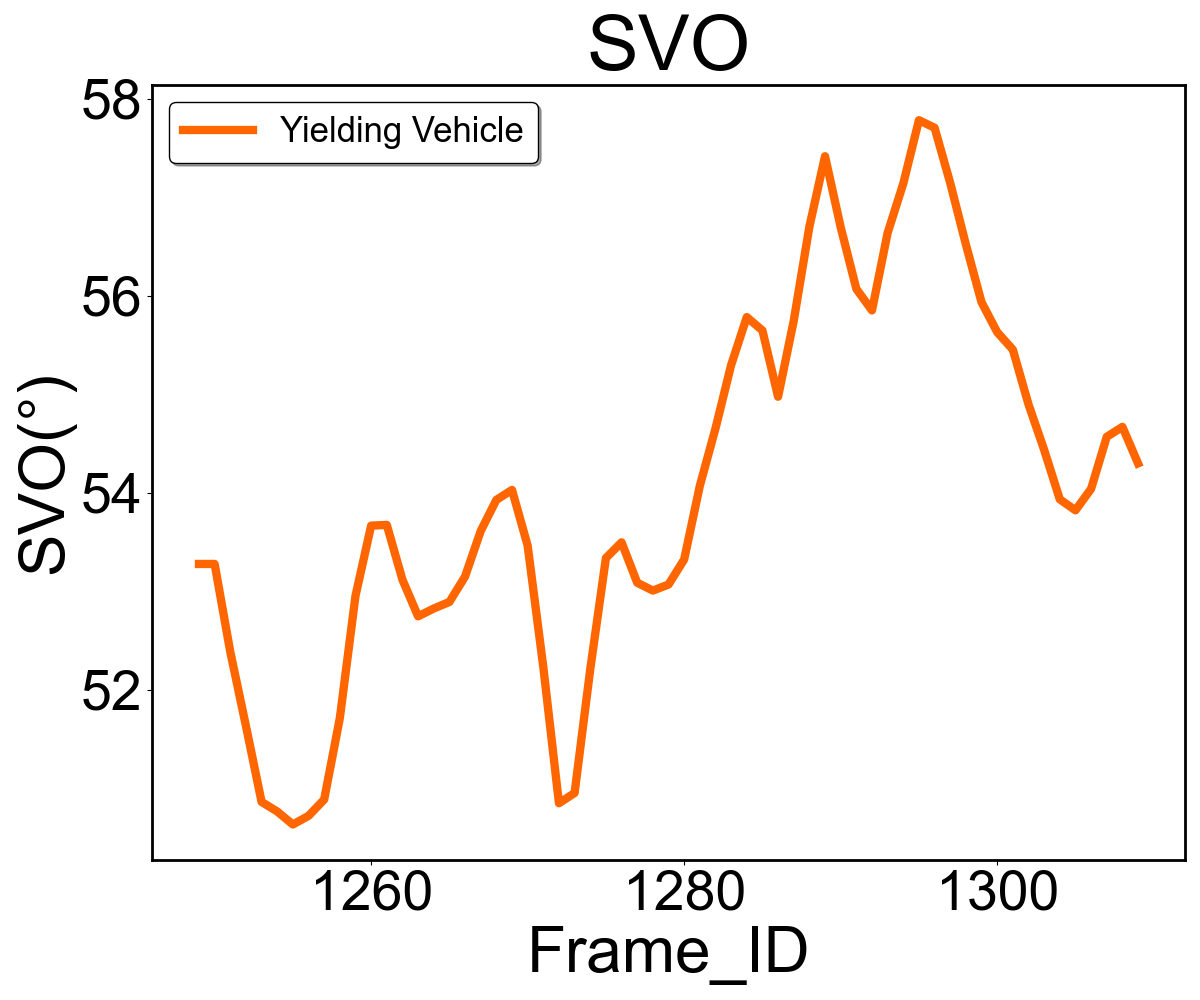} 
    \end{minipage}

    \caption{State Evolution of Vehicles in Yielding Interaction Scenario} 
    \label{fig:all_figures}
\end{figure} 
Figure 8 illustrates the temporal changes in velocity, acceleration, SVO values, and longitudinal coordinates for both the ego vehicle and the conflicting vehicle in this yielding scenario. According to the position figure, it is evident that the ego vehicle's lateral position changes significantly as it transitions from its current lane to the target lane. The lane-change point is distinctly marked at Frame ID 1279, which indicates the precise moment When the ego vehicle is positioned in the middle of the lane-change maneuver, its velocity steadily increases, as indicated by the velocity and acceleration figures. In contrast, the conflicting vehicle's velocity decreases gradually. Meanwhile, the SVO value of the conflicting vehicle continuously increases.

\subsubsection{Passing Scenario}
The dataset includes a total of 20 passing cases and consists of 1220 frames. Each frame represents a sample, capturing the states of the ego vehicle, leading vehicle, and conflicting vehicle.
\begin{figure}[H]
	\centering
	\includegraphics[width=0.77\linewidth]{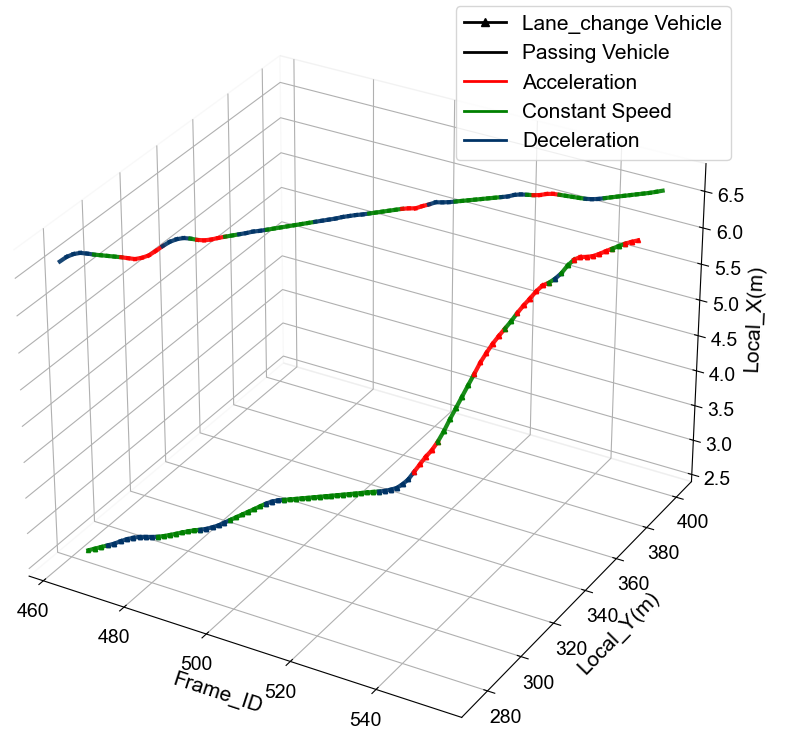}
	\caption{Spatiotemporal Trajectories of Passing Interaction.}
	\label{fig:01}
\end{figure}
Figure 9 and Figure 10 depict one of passing scenarios. Figure 9 provides a three-dimensional visualization of the lane-change process and the conflicting vehicle’s response during the interaction in this scenario. As the ego vehicle transitions to the target lane, the conflicting vehicle chooses to pass it. In this case, the ego vehicle corresponds to vehicle
ID 115, and the conflicting vehicle corresponds to vehicle
ID 116 in the US-101 dataset. 
\begin{figure}[H] 
    \centering 

    \begin{minipage}{0.24\textwidth}
        \centering
        \includegraphics[width=\textwidth]{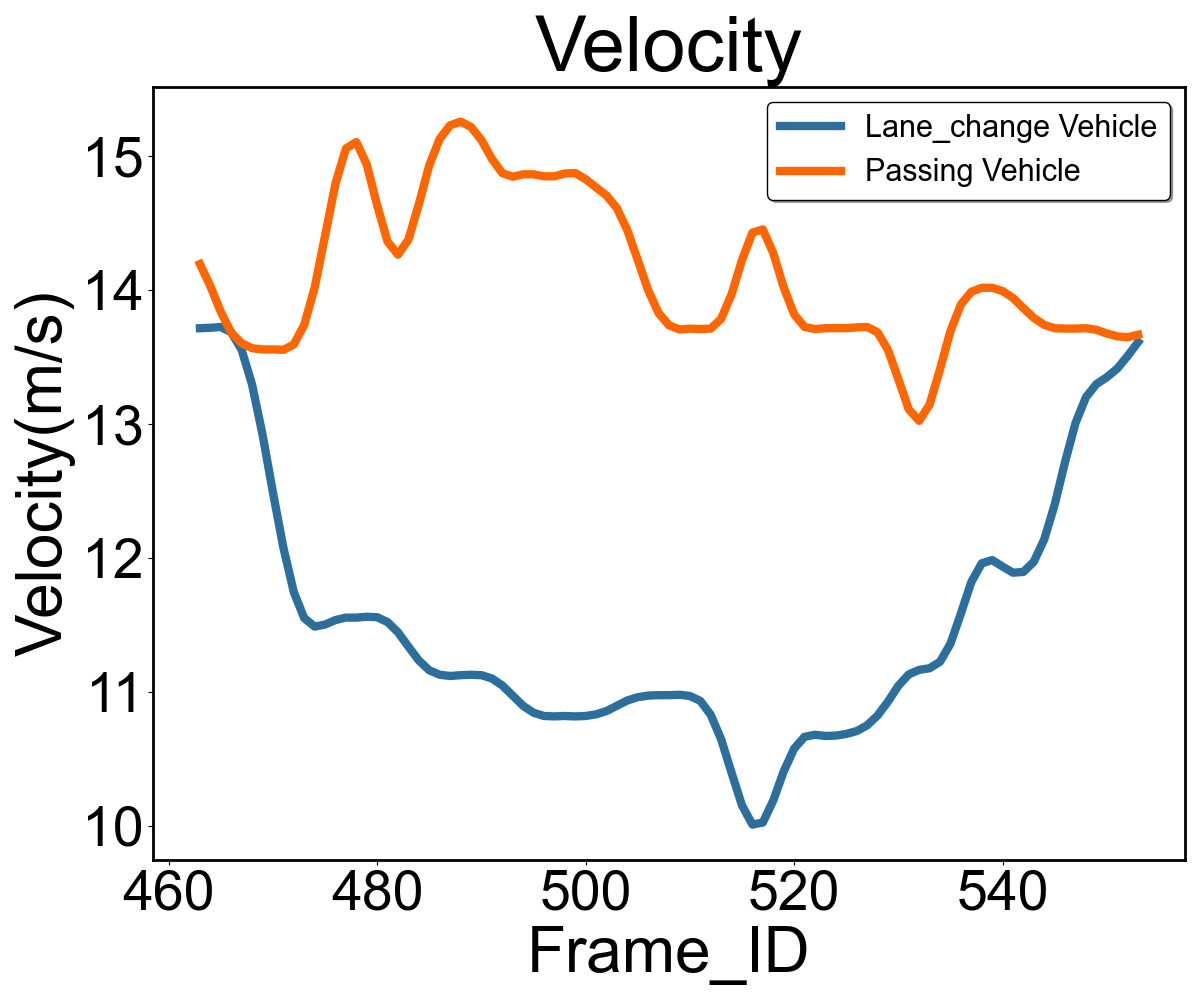} 
    \end{minipage}
    \hfill
    \begin{minipage}{0.24\textwidth}
        \centering
        \includegraphics[width=\textwidth]{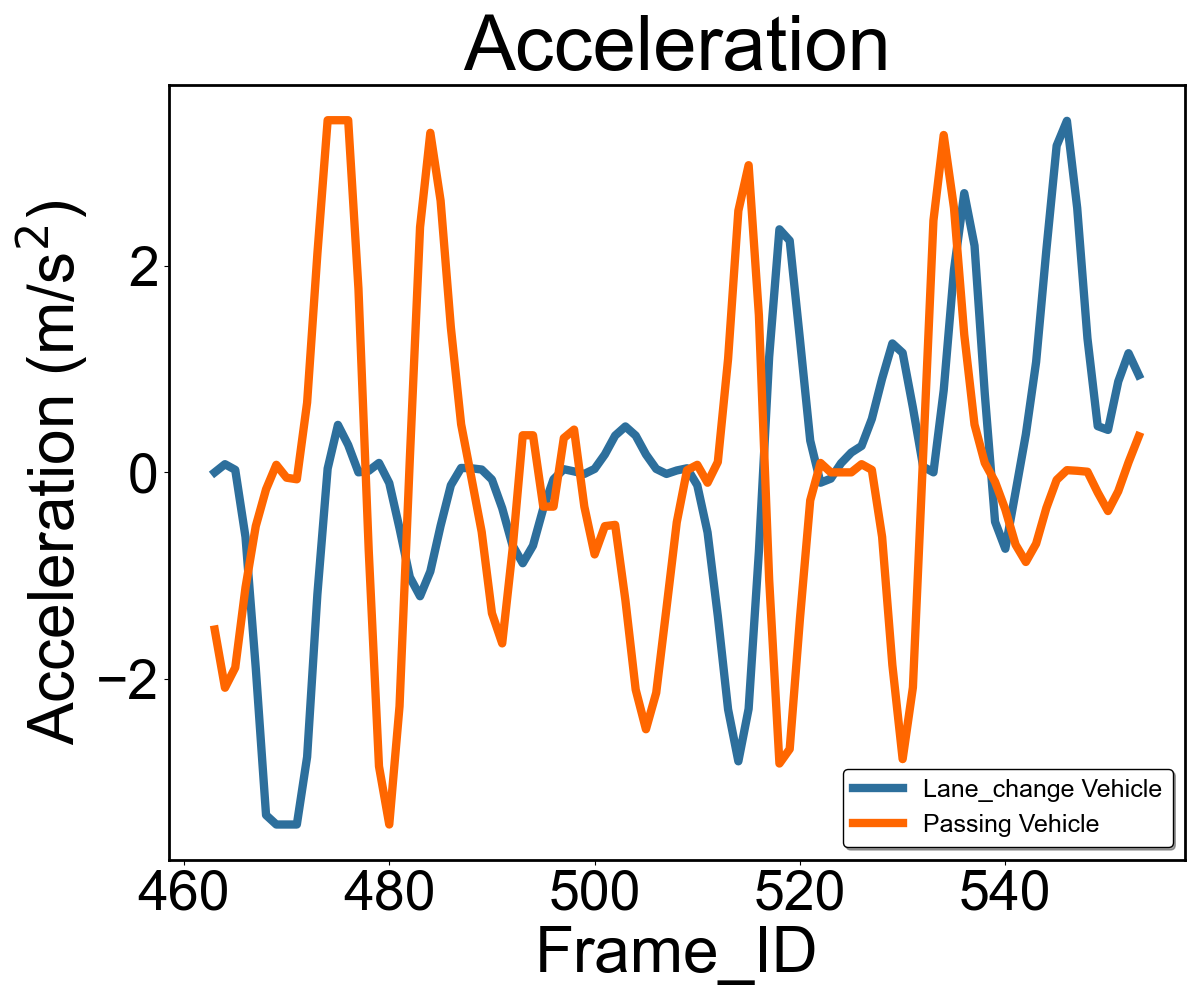} 
    \end{minipage}

    \begin{minipage}{0.24\textwidth}
        \centering
        \includegraphics[width=\textwidth]{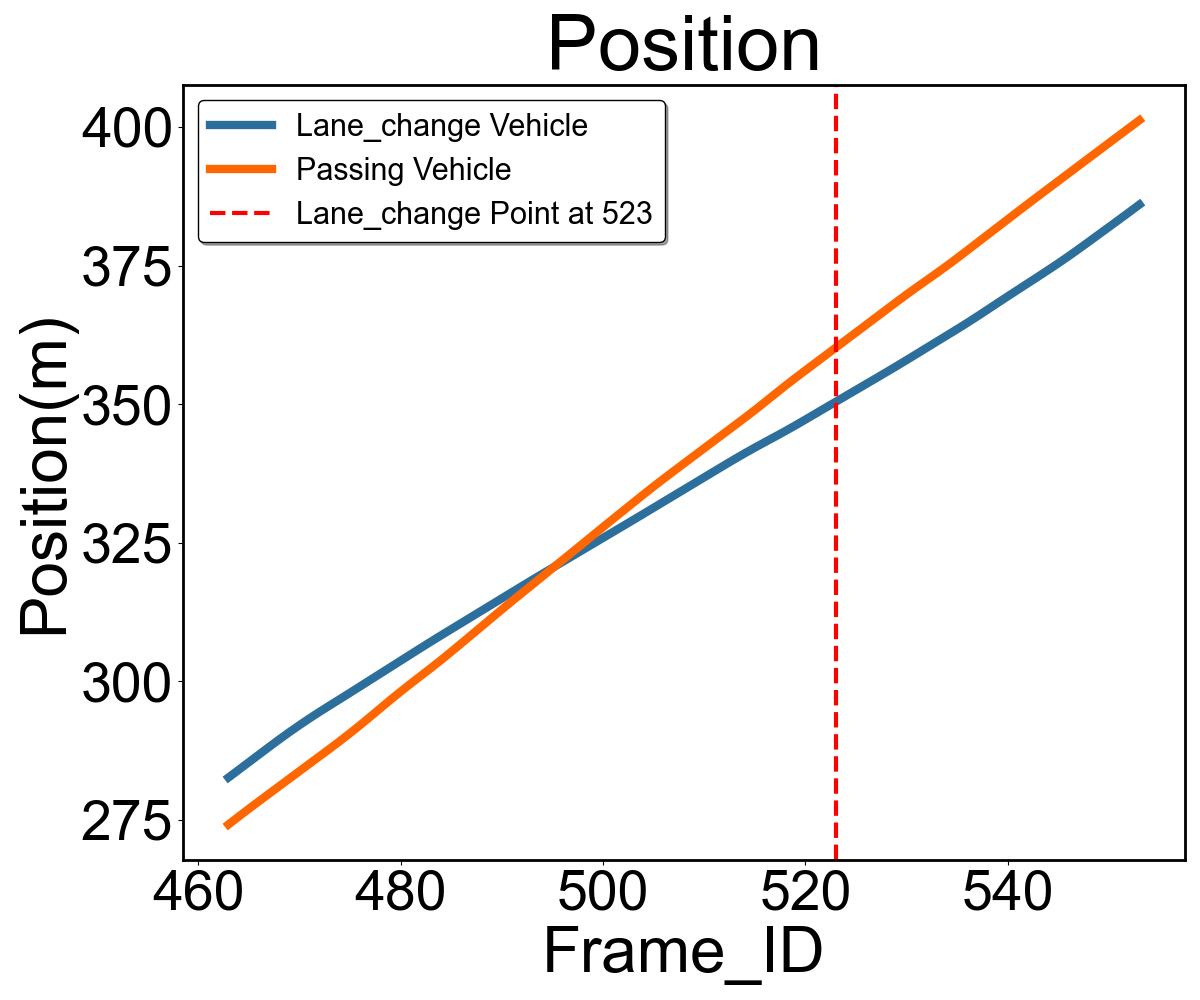} 
    \end{minipage}
    \hfill
    \begin{minipage}{0.24\textwidth}
        \centering
        \includegraphics[width=\textwidth]{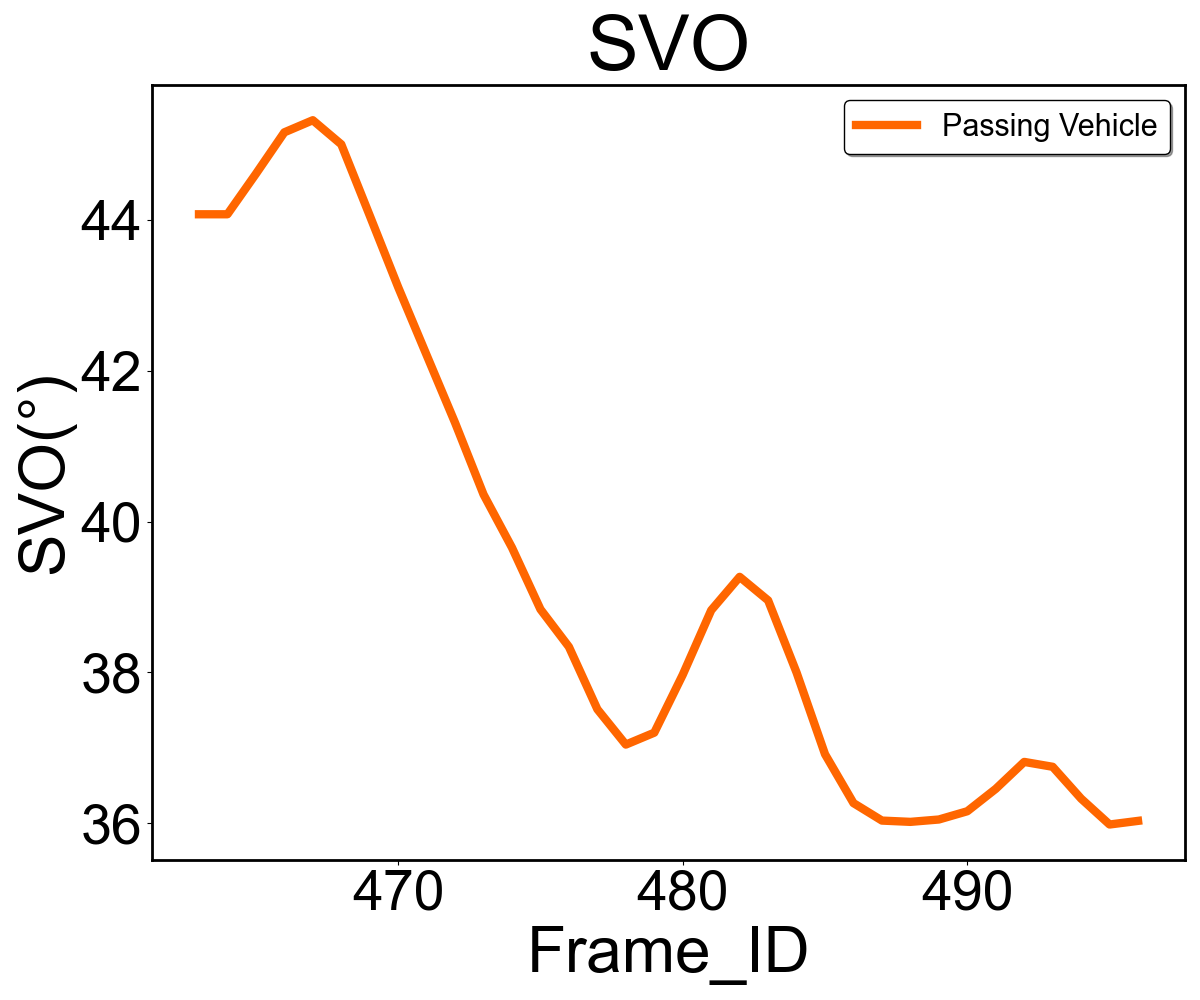} 
    \end{minipage}

    \caption{State Evolution of Vehicles in Passing Interaction Scenario} 
    \label{fig:all_figures}
\end{figure} 
Figure 10 illustrates the temporal variations in velocity, acceleration, SVO values, and longitudinal coordinates for both the ego vehicle and the conflicting vehicle during the passing scenario. The position data clearly indicate the lane-change point is distinctly marked at Frame ID 523, representing the precise moment when the ego vehicle is at the midpoint of the lane-change maneuver. The conflicting vehicle’s velocity steadily increases as it overtakes the ego vehicle, demonstrating a passing behavior. Meanwhile, the ego vehicle’s velocity gradually decreases as the conflicting vehicle completes the overtaking maneuver. During the overtaking process, the SVO of the conflicting vehicle decreases.
\begin{figure}[H] 
    \centering 
    \begin{minipage}{0.241\textwidth}
        \centering
        \includegraphics[width=\textwidth]{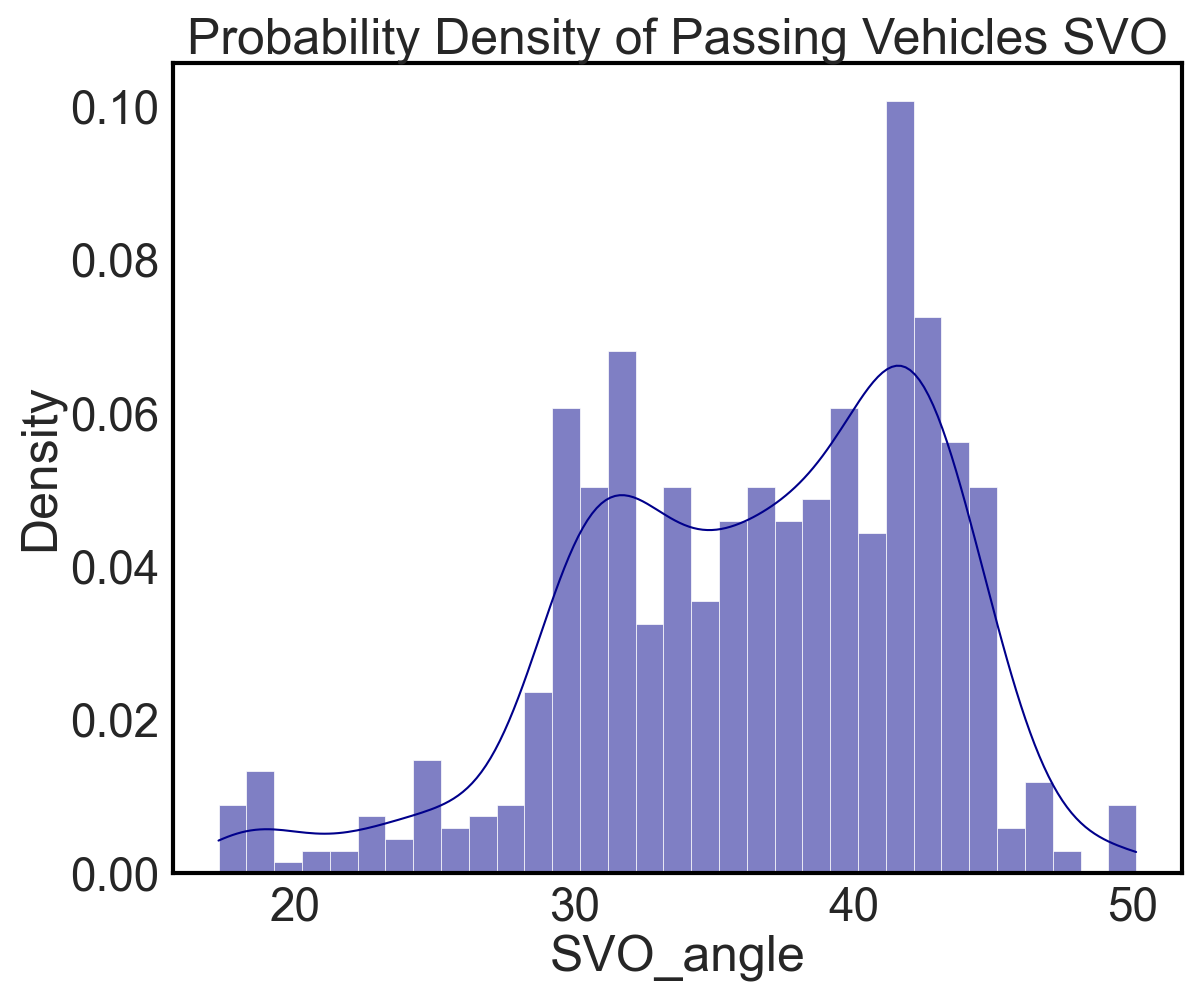} 
    \end{minipage}
    \hfill
    \begin{minipage}{0.241\textwidth}
        \centering
        \includegraphics[width=\textwidth]{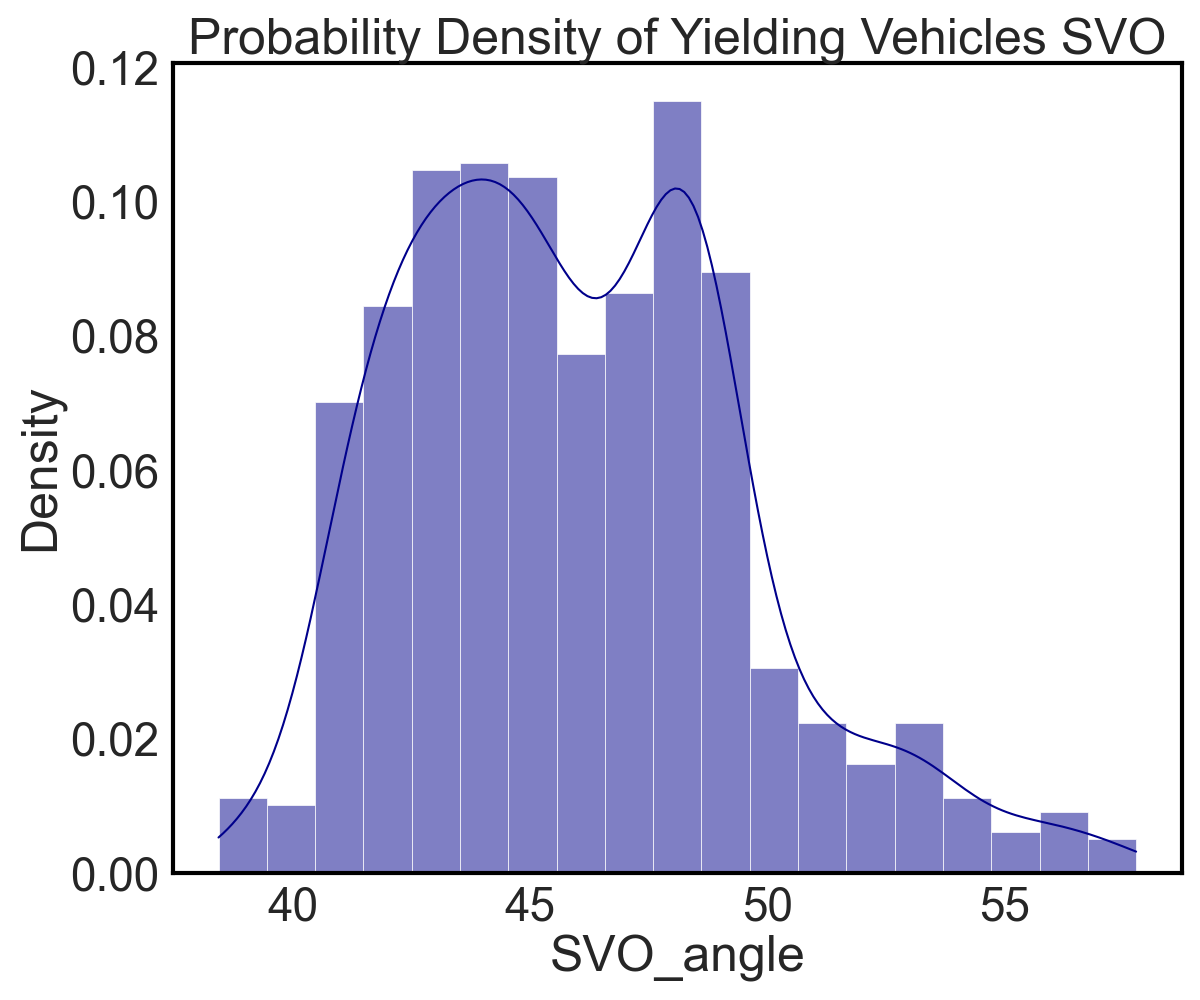} 
    \end{minipage}
    \caption{Probability Density Distribution of SVO Angles in Passing and Yielding Scenarios.} 
    \label{fig:all_figures}
\end{figure} 
It is crucial to understand how individuals’ social preferences can influence their behaviors in lane-changing scenarios by analyzing the distribution of SVO for both yielding and passing intents as illustrated in Figure 11. Specifically, within all of yielding cases the conflicting vehicles exhibit more prosocial orientations, characterized by a higher concern for the interests of others, are more likely to yield to the ego vehicles during lane changes in the left side of Figure 11. Conversely, the conflicting vehicles with a more individualistic orientation, characterized by a higher focus on their own efficiency, being more inclined to pass to the ego vehicles rather than yielding. Therefore, this behavior reflects a prioritization of one’s own objectives over considerations for others’ needs or safety.

\section{Simulation setup}
\subsection{Driving Scenario Setting}
To evaluate the effectiveness of the proposed decision-making algorithm for autonomous vehicles, a highway driving scenario is built using the OpenAI Gym-based highway-env simulator \cite{highway-env}, which provides flexible configuration of traffic density and lane numbers within the environment. The constructed highway scenario has three lanes where the AV to be controlled was surrounded by 2 other vehicles, as illustrated in Figure 2. In the simulator, the dimensions of all vehicles are set with a length of 5 meters and a width of 2 meters. In this specific task, the primary objective for the AV is to perform a lane-change behavior.
\subsection{Surrounding Vehicles' Behavior Model}
The behavior of surrounding vehicles is simulated using two widely adopted microscopic traffic models: the Intelligent Driver Model (IDM) \cite{Treiber2000CongestedTS} for longitudinal control and the Minimizing Overall Braking Induced by Lane changes (MOBIL) model \cite{Kesting2007MOBILG} for lateral movements. These models collectively facilitate the generation of realistic vehicle interactions in highway scenarios.\\
In terms of IDM model, it is a car-following model, simulates how vehicles adjust their speed based on the car ahead. It considers factors like desired velocity \(v_0\), the distance between the vehicle and the leading vehicle \(s\), the maximum acceleration of the vehicle \(a_{\text{max}}\), and the speed of the vehicle \(v\) to replicate realistic driving behaviors.
\begin{equation}
    a = a_{\text{max}} \left[1 - \left(\frac{v}{v_0}\right)^\delta - \left(\frac{s^*(v,\Delta v)}{s}\right)^2\right]
\end{equation}
where \(s^*(v,\Delta v)\) is the safety distance function, which depends on the current speed \(vT\) and the speed difference \(\Delta v\).
\begin{equation}
    s^*(v,\Delta v) = s_0 + vT + \frac{v\Delta v}{2\sqrt{a_{\text{max}}b}}
\end{equation}
MOBIL model focuses on lane-changing decisions. It evaluates when a driver should switch lanes by weighing the benefits against potential risks. The model takes into account the impact on surrounding vehicles, aiming to minimize overall braking in the traffic flow.
\begin{equation}
\begin{aligned}
    \tilde{a}_n &> -b_{\text{safe}} \\[10pt]
\tilde{a}_c - a_c &> p(a_n + \tilde{a}_o - a_o) + a_{\text{thr}}
\end{aligned}
\end{equation}
where \(\tilde{a}_c \) is the acceleration of the changing vehicle after the change, \(a_c\) is the current acceleration of the changing vehicle, p is the politeness factor, \(a_n\) is the new follower's acceleration, \(\tilde{a}_o\) is the old follower's acceleration after the change, \(a_o\) is the current acceleration of the old follower, and \(b_{\text{safe}}\) is the safe braking threshold.

\subsection{Implementation details}
\subsubsection{Training details}
To train an optimal behavioral policy for the AV, the proposed decision-making algorithm DQN-YI is implemented within a PyTorch-based framework and trained across 5000 simulation episodes.Each episode terminates either upon the AV successfully reaching its destination or following a collision with surrounding traffic participants. Additionally, the reward values for each episode are recorded, facilitating model evaluation and testing. Table I outlines the selected hyperparameters for the developed algorithm.
\begin{table}[H]\tiny
  \centering
  {\fontsize{10pt}{10pt}\selectfont
  \renewcommand{\arraystretch}{1.3}
  \setlength{\tabcolsep}{15pt}
  \caption{Hyperparameter used for training.}
  \label{table:hyperparameters}
  \begin{tabular}{@{\hspace{1em}} l c r @{\hspace{1em}}} \hline
Discounted factor     & \(\gamma\)                   & 0.9  \\ 
Replay memory size    & \( M_{\text{replay}} \)      & 2,000    \\ 
Mini-batch size       & \( M_{\text{mini}} \)        & 32      \\ 
Learning rate & \( \eta \) & 0.001 \\ 
Epsilon & \(\epsilon\) & 0.1 \\ 
Target update frequency & $N_{\text{update}}$ & 1000 \\ \hline
  \end{tabular}
  }
\end{table}

\subsubsection{Performance Metrics}
To evaluate and validate the performance of our algorithm, we conduct tests on each trained policy and record several key metrics, including collision rate, successful rate, average speed, and average reward.
\begin{itemize}
\item[$\bullet$] Collision Rate: The metric quantifies the frequency at which collisions occur between the AV and surrounding vehicles during the testing period.
\end{itemize}
\begin{itemize}
\item[$\bullet$] Successful Rate: This metric measures the success rate of lane change completions over the evaluation period.
\end{itemize}
\begin{itemize}
\item[$\bullet$] Average Speed: The metric represents the mean velocity of the AV over the course of the testing period. This metric serves as an indicator of the vehicle's overall performance in terms of its speed management and efficiency within the simulated driving scenario.
\end{itemize}
\begin{itemize}
\item[$\bullet$] Average Reward: This metric represents the mean of the cumulative rewards earned by the AV during the testing period.
\end{itemize}

\subsubsection{Comparison Baselines}
The classical DQN, Dueling DQN, Double DQN, Advantage Actor-Critic (A2C), and Dueling Double DQN (D3QN) algorithms has been selected as baseline methods for comparison. To ensure consistency in the comparison, these baseline algorithms are configured with identical hyperparameters to those of the developed algorithm.\\
DQN: DQN leverages deep neural networks to approximate the Q-function, enabling effective action-value estimation in complex state spaces. \\
Double DQN:Double DQN extends the original DQN framework by mitigating the overestimation bias commonly observed in Q-value predictions. It retains the same network architecture, input observations, and output control mechanisms as those employed in the proposed method.\\
Dueling DQN: The Dueling architecture enhances the DQN algorithm by decoupling the estimation of the state value from the advantage associated with each action, thereby improving learning efficiency. \\
D3QN: D3QN is an advanced reinforcement learning algorithm that uses the dueling network architecture to provide a more nuanced understanding of the state and action values while employing the double DQN methodology to mitigate value overestimations.\\
A2C: A2C is a reinforcement learning algorithm that integrates value-based and policy-based strategies to improve the stability and efficiency of the learning process. It achieves this by concurrently updating two distinct components: a policy network (actor) responsible for action selection, and a value network (critic) that estimates the expected returns.

\section{Results and Discussion}
\begin{table*}[htbp]
\centering
\setlength{\tabcolsep}{20pt}
\caption{Policy Evaluation Results}
\renewcommand{\arraystretch}{1.3}
\begin{tabularx}{\textwidth}{ccccc}
\toprule
\bfseries Models &\bfseries Collision Rate (\%) &\bfseries Successful Rate (\%) &\bfseries Average Reward &\bfseries Average Speed (m/s) \\
\midrule
A2C   & 20.98 &78.16 & 6.69 & 25.82  \\
DQN   & 16.74 & 83.10 & 7.87 & 26.58  \\
D3QN       & 16.22  & 83.34  &8.21 & 26.41  \\
Double DQN       & 15.96 & 83.82 & 8.65 & 24.91  \\
Dueling DQN      & 14.84  & 84.76 & 8.83 & 26.92  \\
\bfseries DQN-YI (Ours) & \bfseries 7.52 & \bfseries 92.02 & \bfseries 10.63 & \bfseries 26.20 \\
\bottomrule
\end{tabularx}
\end{table*}

After completing the above necessary setup, this section includes the initiation of the simulation and the execution of the training loop for the specified number of iterations. Throughout this process, episodic rewards are recorded to evaluate the agent’s performance in lane-changing maneuvers, along with other relevant evaluation metrics. Furthermore, a brief explanation of the results is provided, highlighting key characteristics and insights derived from the analysis.
\subsection{Policy Convergence}
\begin{figure}[H]
	\centering
	\includegraphics[width=0.9\linewidth]{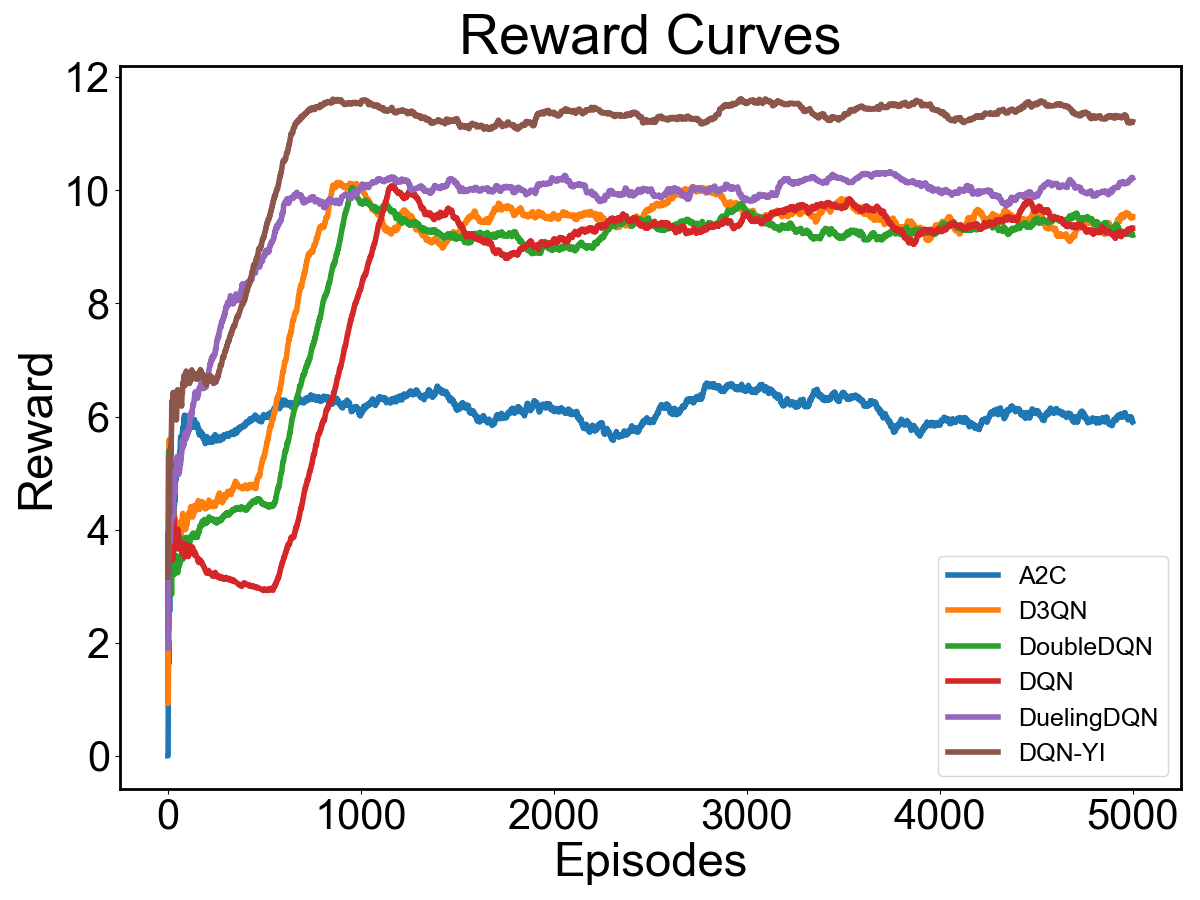}
	\caption{The reward curves.}
	\label{fig:01}
\end{figure}
This subsection outlines the training procedure for our proposed decision-making algorithm, which aims for AV to maximize the cumulative reward in a single episode. An increase in cumulative reward during training indicates policy improvement, while convergence signifies that the policy has reached its local maximum reward. The agents are trained using the hyper-parameters specified in Table I. 
Figure 12 illustrates the training process of DQN-YI compared to other baseline methods, demonstrating that the average cumulative rewards achieved by DQN-YI consistently increase and eventually surpass those of the baseline methods.
\subsection{Performance Evaluation}
 To assess the performance of the trained DQN-YI agent, it is essential to conduct testing and validation. This ensures the agent can consistently handle unexpected situations and safely interact with other surrounding vehicles during lane changes. The test scenario in this subsection is identical to the training scenario. We evaluate the policy of the DQN-YI algorithm over 5000 episodes in this setting using several key metrics and compare its performance with that of baseline algorithms.
\subsubsection{Qualitative Results}
It is significant to interpret the learned lane-change behavior of the AV in the test scenario, as shown by the results depicted in Figures 13 and 14.\\
\textit{\textbf{Case 1:}} This case, as shown in Figure 13, illustrates the AV’s lane-change maneuver in response to a yielding vehicle. Initially, the AV is positioned at 80 meters longitudinally and 6 meters laterally, traveling at 30 m/s. The leading vehicle located at 130 meters longitudinally and 6 meters laterally is traveling at a constant speed of 25 m/s in the same lane. Meanwhile, the yielding vehicle in the adjacent lane is initially positioned 10 meters behind the AV longitudinally, traveling at 42 m/s. Throughout the interaction, the AV maintains its velocity while inferring the yielding vehicle's intention before initiating the lane change.
 \begin{figure}[htbp] 
    \centering 

    \begin{minipage}{0.152\textwidth} 
        \centering
        \includegraphics[width=\textwidth]{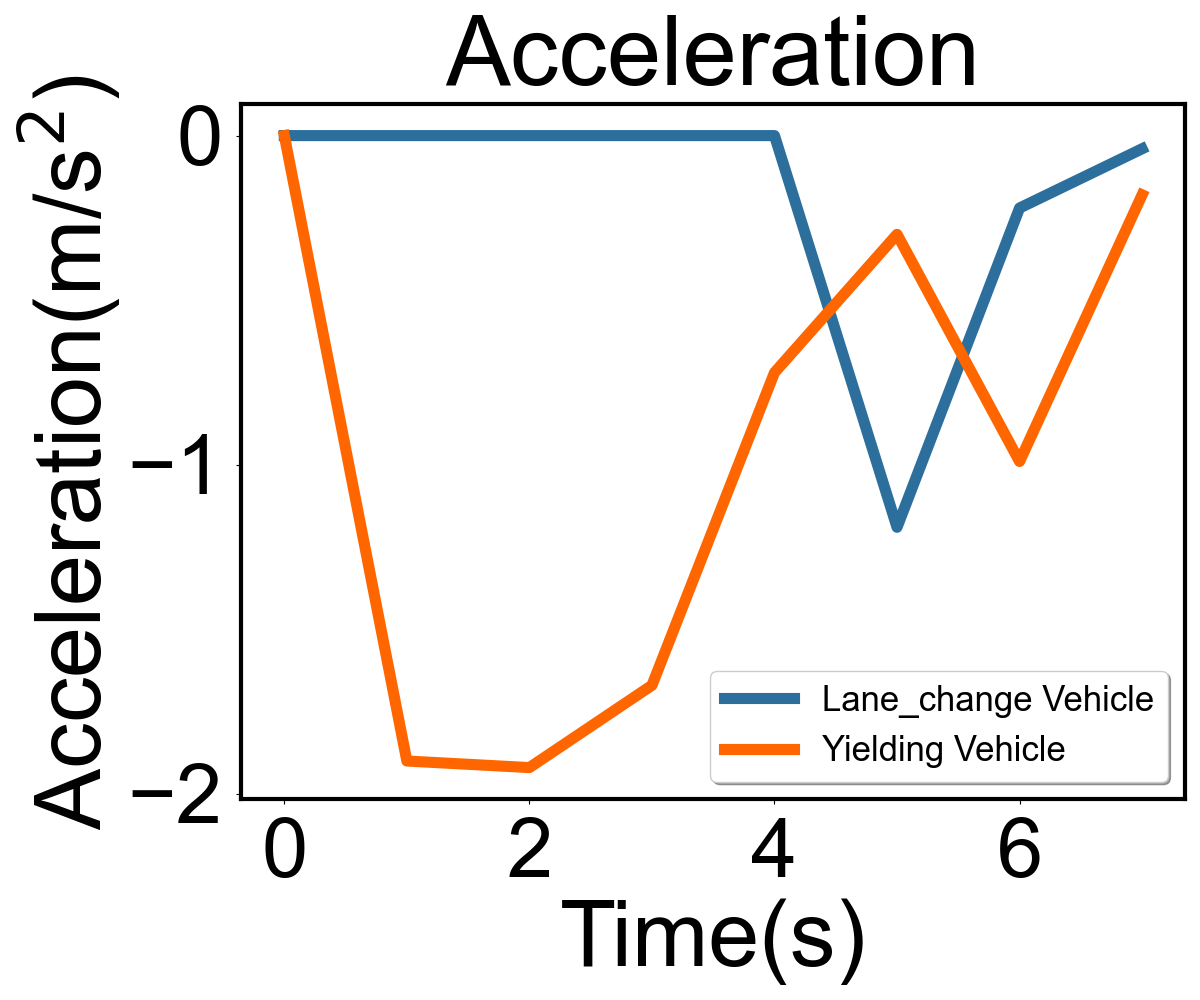} 
        \label{fig:vel}
    \end{minipage}
    \hfill
    \begin{minipage}{0.152\textwidth}
        \centering
        \includegraphics[width=\textwidth]{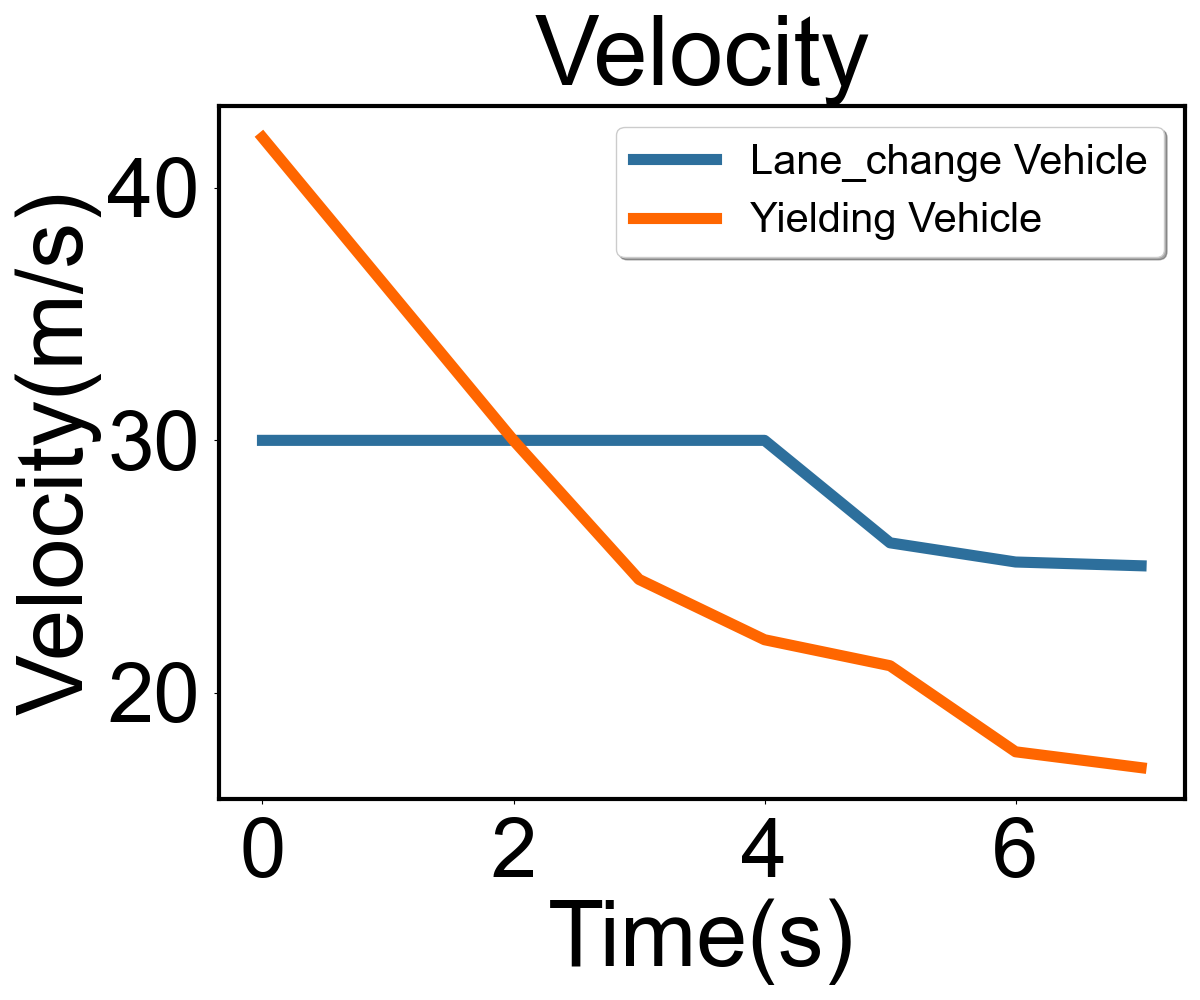} 
        \label{fig:acc}
    \end{minipage}
    \hfill
    \begin{minipage}{0.152\textwidth}
        \centering
        \includegraphics[width=\textwidth]{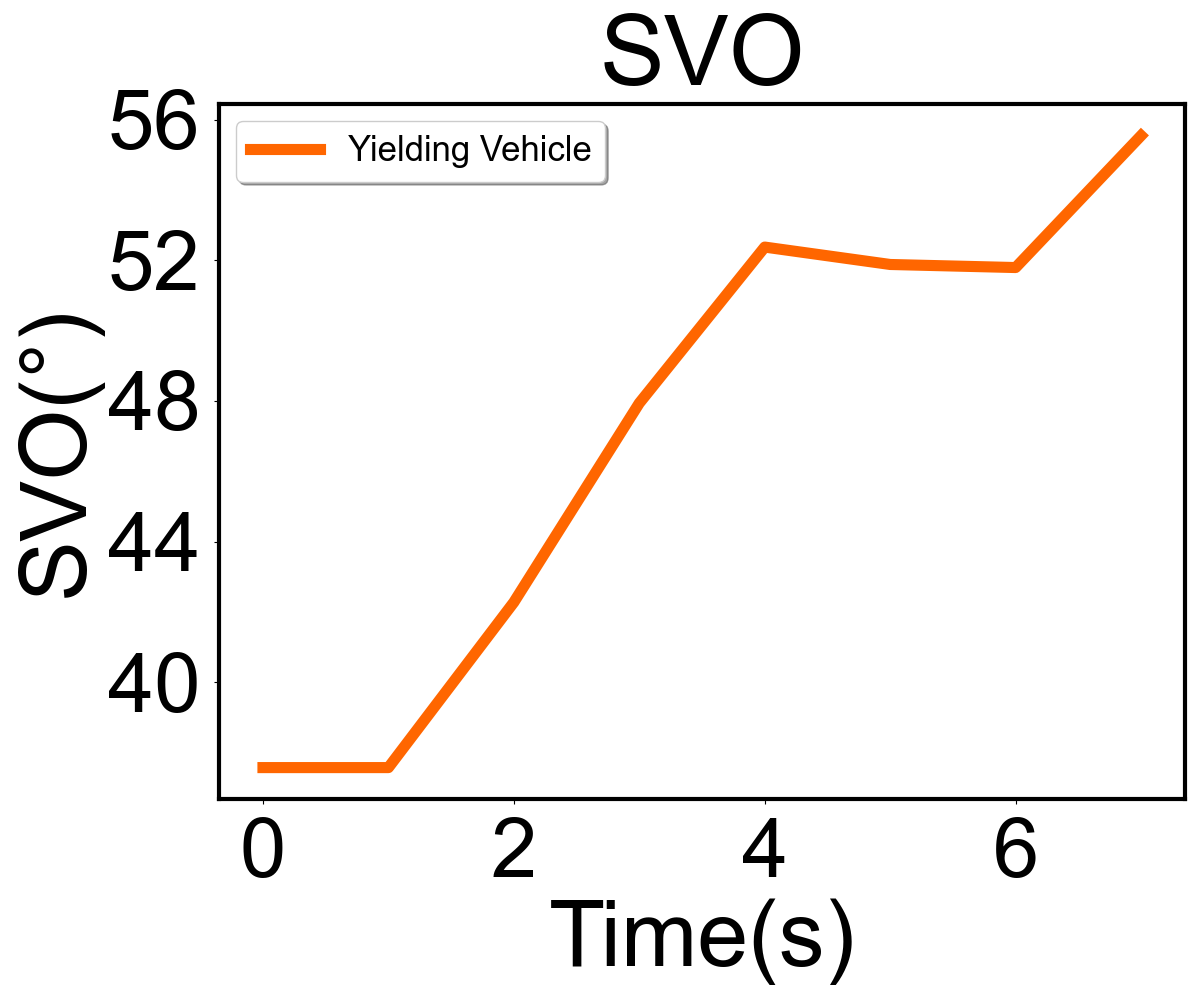} 
        \label{fig:pos}
    \end{minipage}

    \begin{minipage}{0.153\textwidth}
        \centering
        \includegraphics[width=\textwidth]{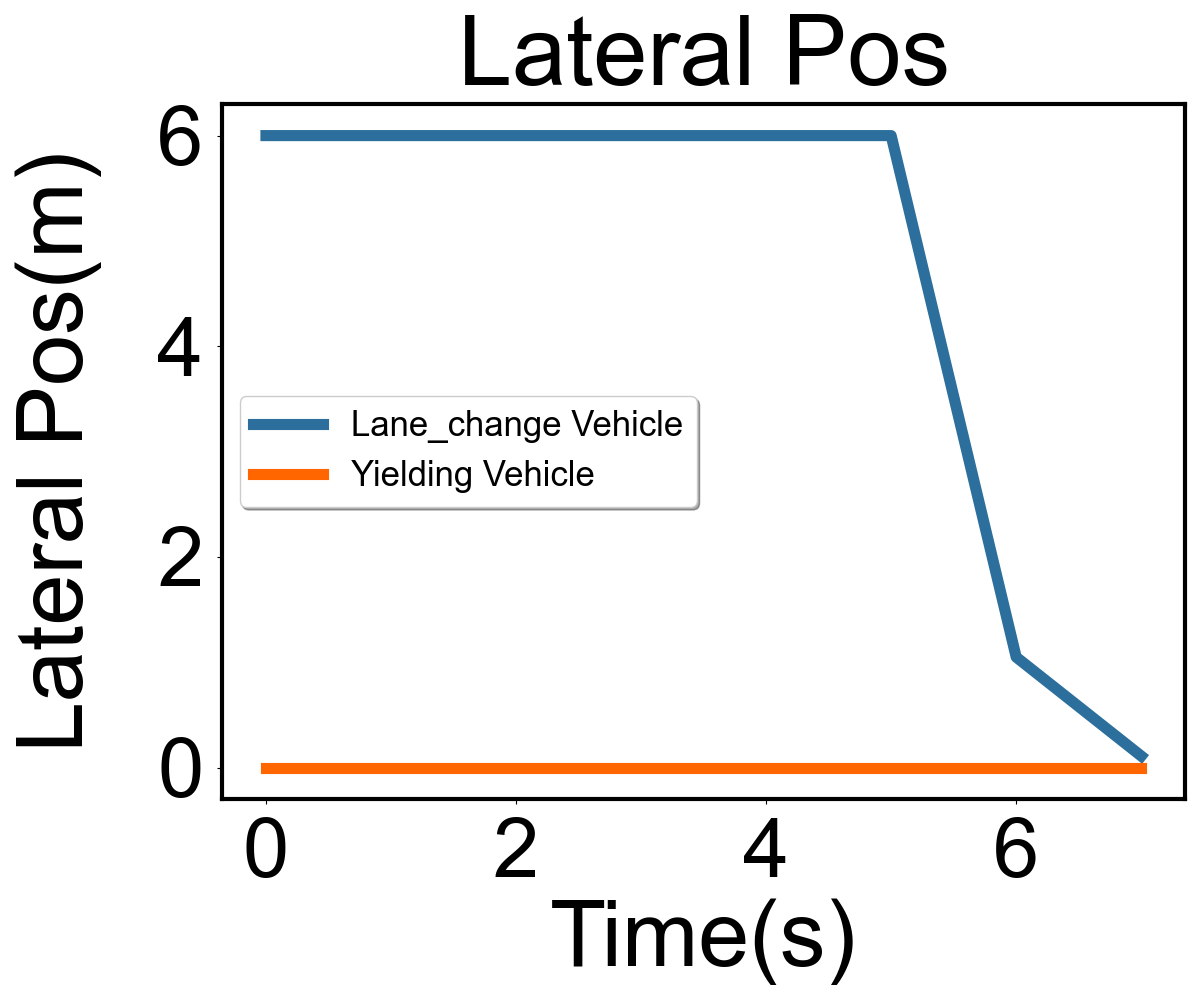} 
        \label{fig:svo}
    \end{minipage}
    \hfill
    \begin{minipage}{0.155\textwidth}
        \centering
        \includegraphics[width=\textwidth]{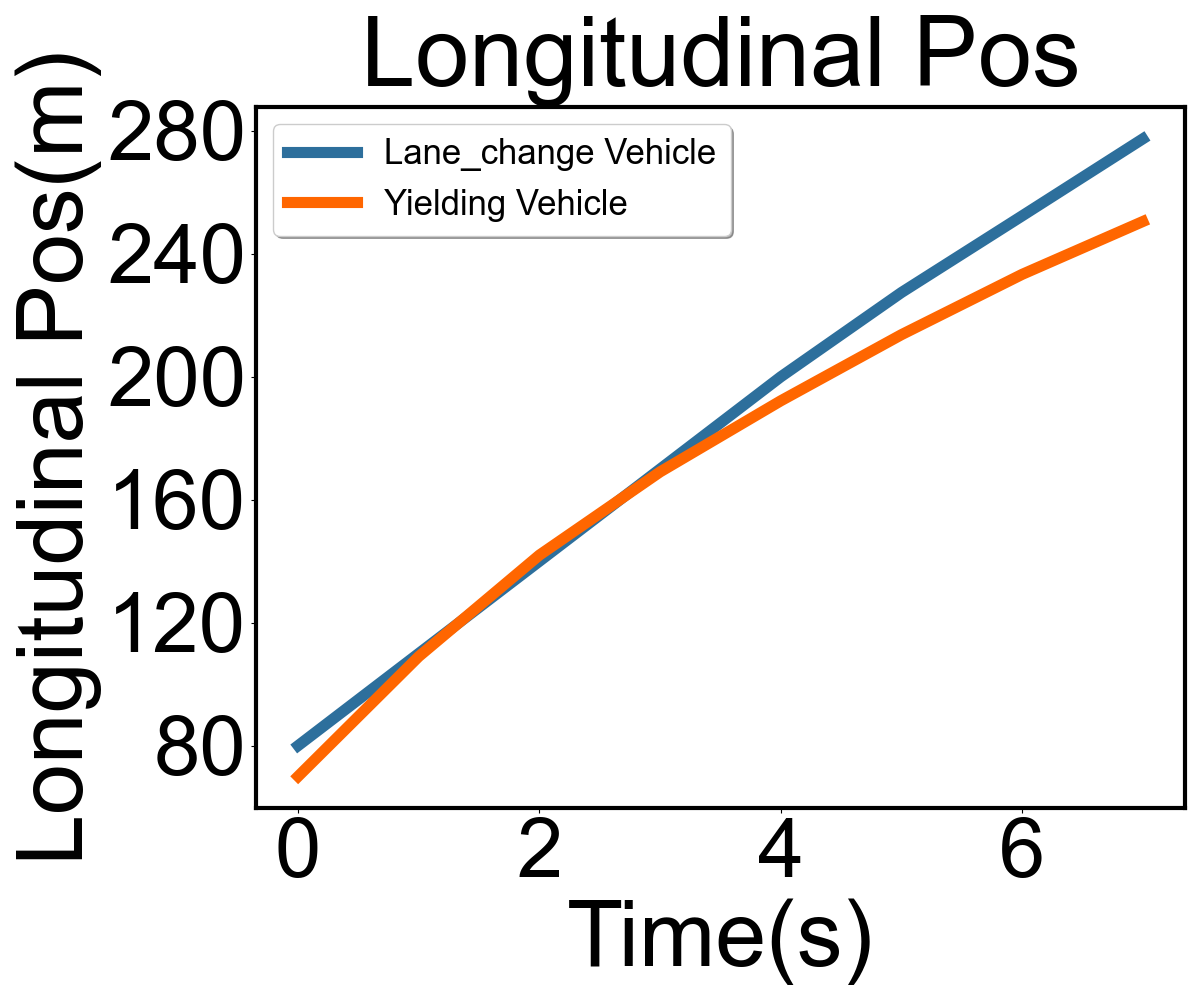} 
        \label{fig:acc}
    \end{minipage}
    \hfill
    \begin{minipage}{0.155\textwidth}
        \centering
        \includegraphics[width=\textwidth]{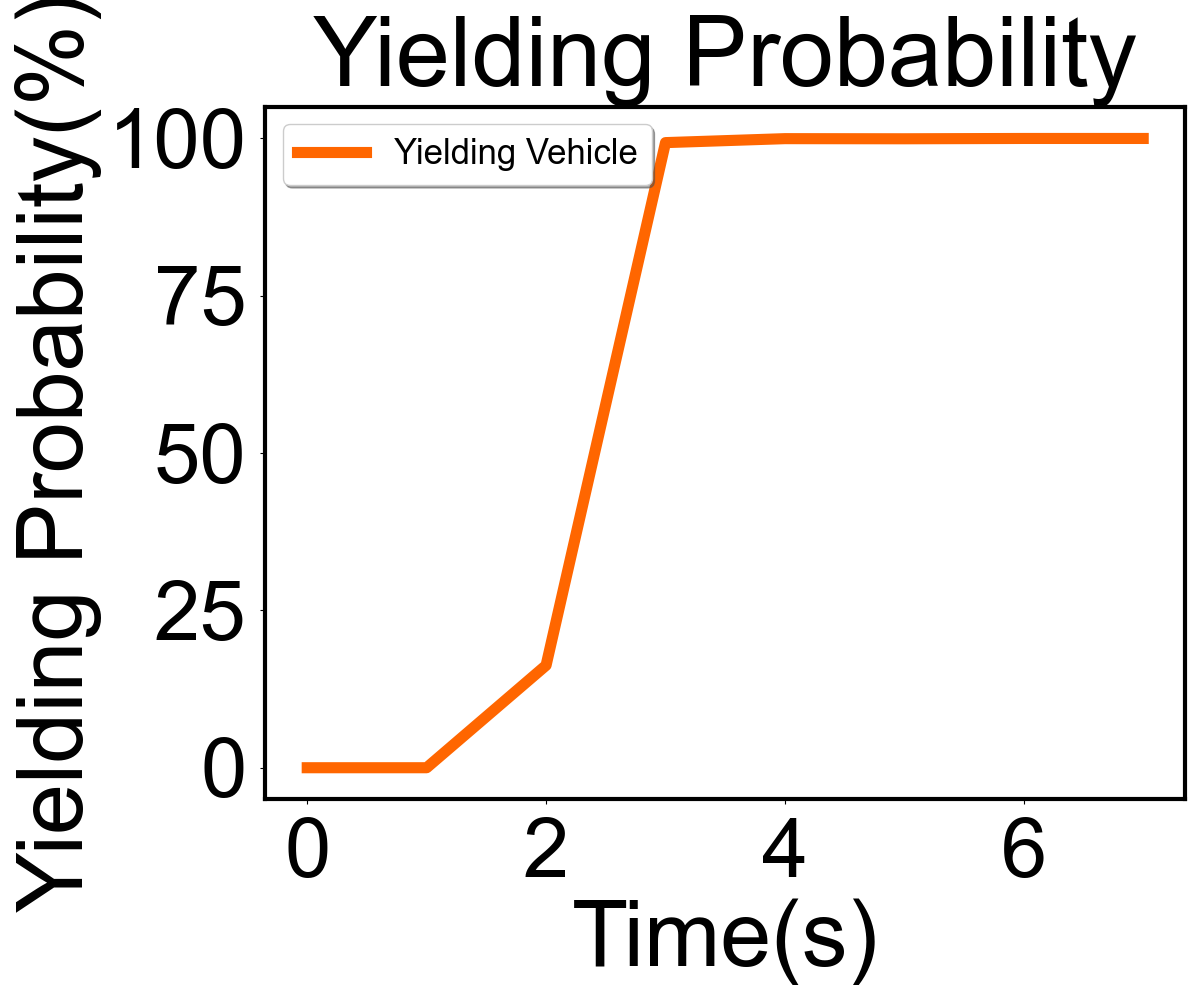} 
        \label{fig:pos}
    \end{minipage}

    \caption{Temporal Evolution of Vehicle States in Case 1.} 
    \label{fig:all_figures}
\end{figure}
Specifically, it reveals that as the AV executes its lane-changing maneuver and approaches its target position, the yielding vehicle demonstrates yielding behavior by gradually reducing its velocity, as evidenced by the acceleration and velocity profiles. The SVO analysis further indicates the yielding vehicle's increasing willingness to yield, shown by its decreasing consideration of self-interest. The computed yielding probability progressively rises during this interaction. Thus, this process successfully demonstrates the AV's capability to perform safe lane changes while accurately inferring and responding to the surrounding vehicle's yielding behavior.\\
\textit{\textbf{Case 2:}} The lane-change sequence of the AV when a passing intention is displayed in Case 2 is illustrated in Figure 14. Initially, the AV is positioned at 80 meters longitudinally and 6 meters laterally, traveling at 30 m/s. The leading vehicle of the AV, located at 130 meters longitudinally and 6 meters laterally, is moving at a constant velocity of 25 m/s in the same lane. Meanwhile, the passing vehicle in the adjacent lane is initially positioned 10 meters behind the AV longitudinally, traveling at 40 m/s.
\begin{figure}[htbp] 
    \centering 
    \begin{minipage}{0.153\textwidth} 
        \centering
        \includegraphics[width=\textwidth]{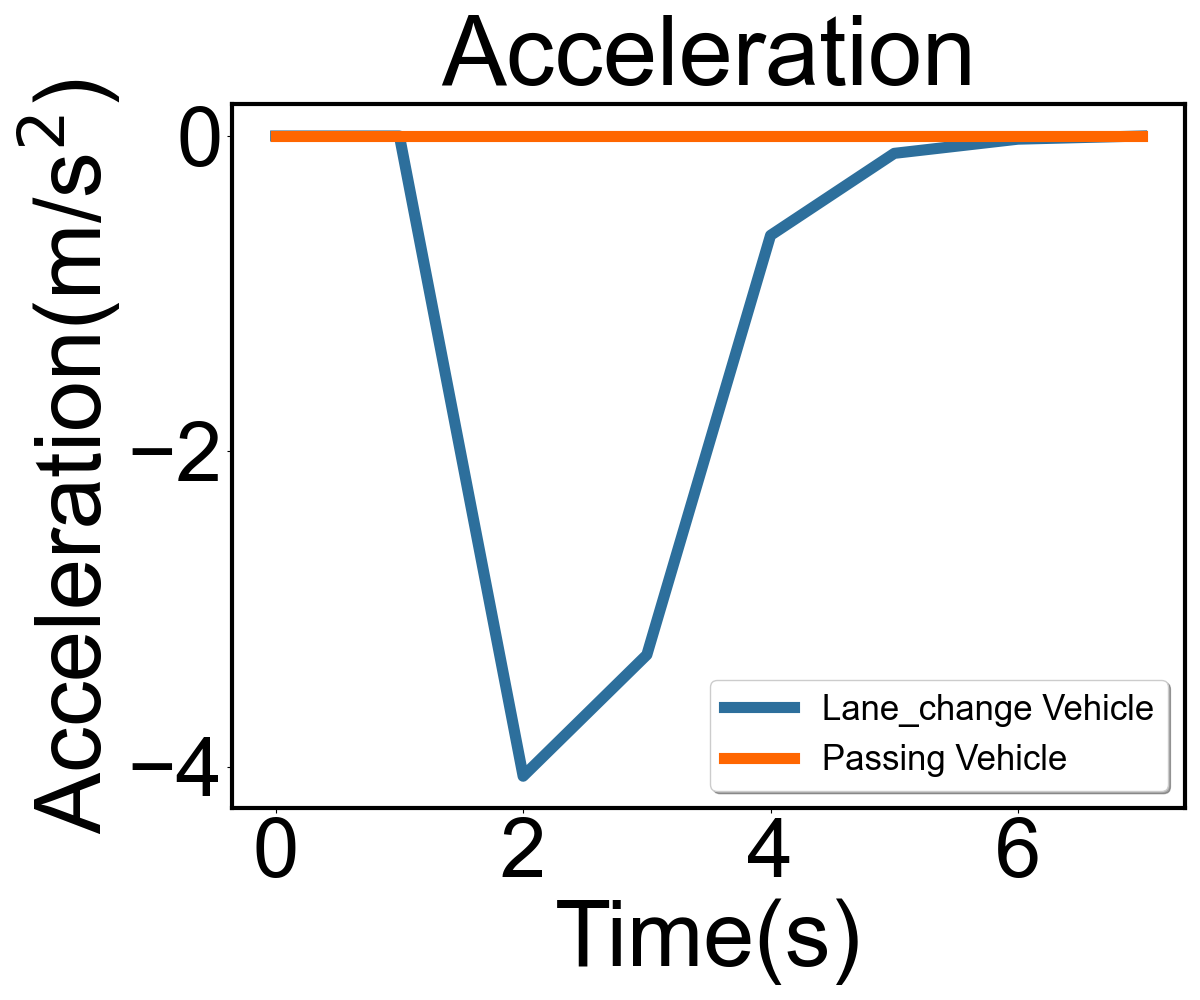} 
        \label{fig:vel}
    \end{minipage}
    \hfill
    \begin{minipage}{0.155\textwidth}
        \centering
        \includegraphics[width=\textwidth]{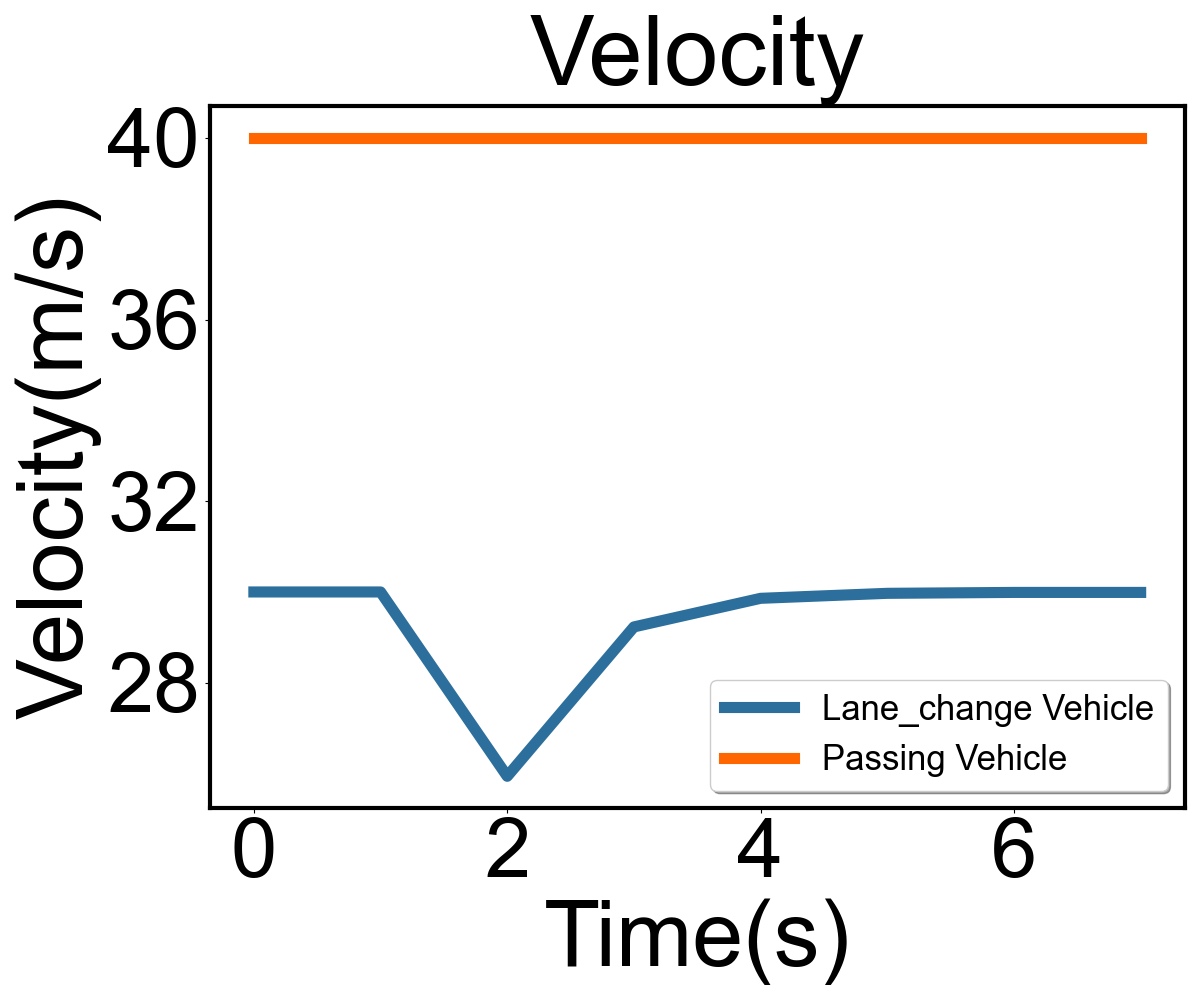} 
        \label{fig:acc}
    \end{minipage}
    \hfill
    \begin{minipage}{0.155\textwidth}
        \centering
        \includegraphics[width=\textwidth]{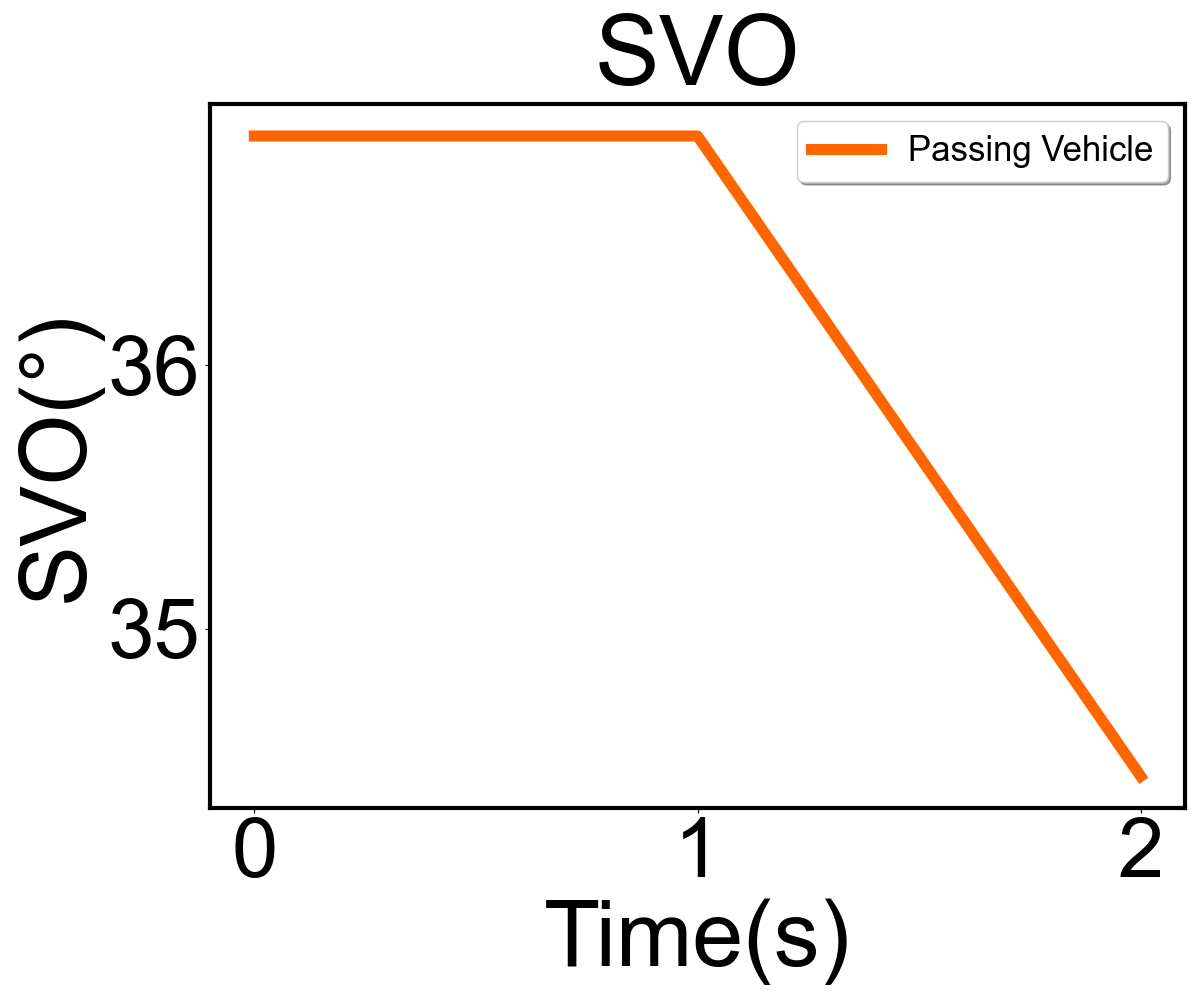} 
        \label{fig:pos}
    \end{minipage}
    \begin{minipage}{0.153\textwidth}
        \centering
        \includegraphics[width=\textwidth]{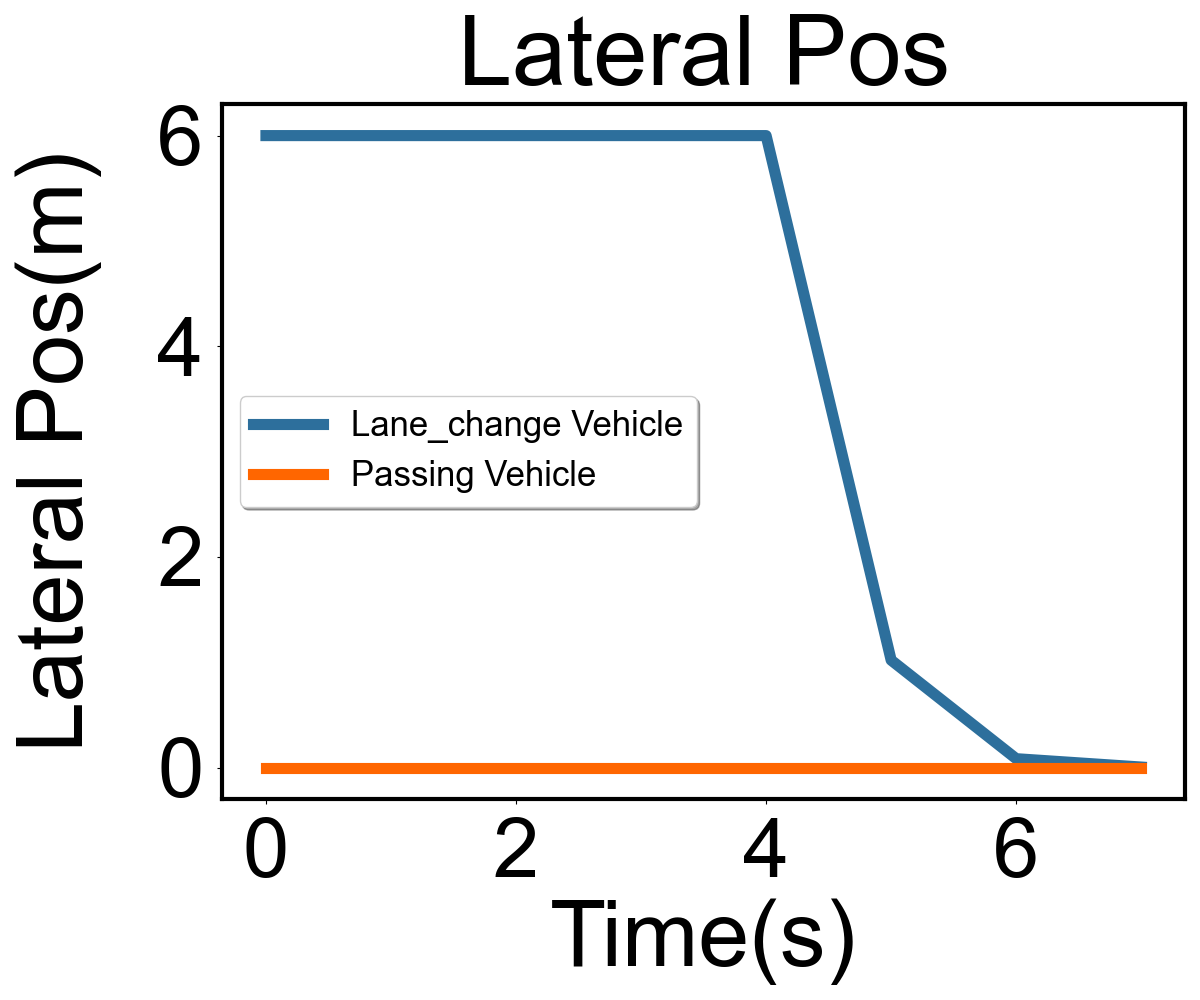} 
        \label{fig:svo}
    \end{minipage}
    \hfill
    \begin{minipage}{0.157\textwidth}
        \centering
        \includegraphics[width=\textwidth]{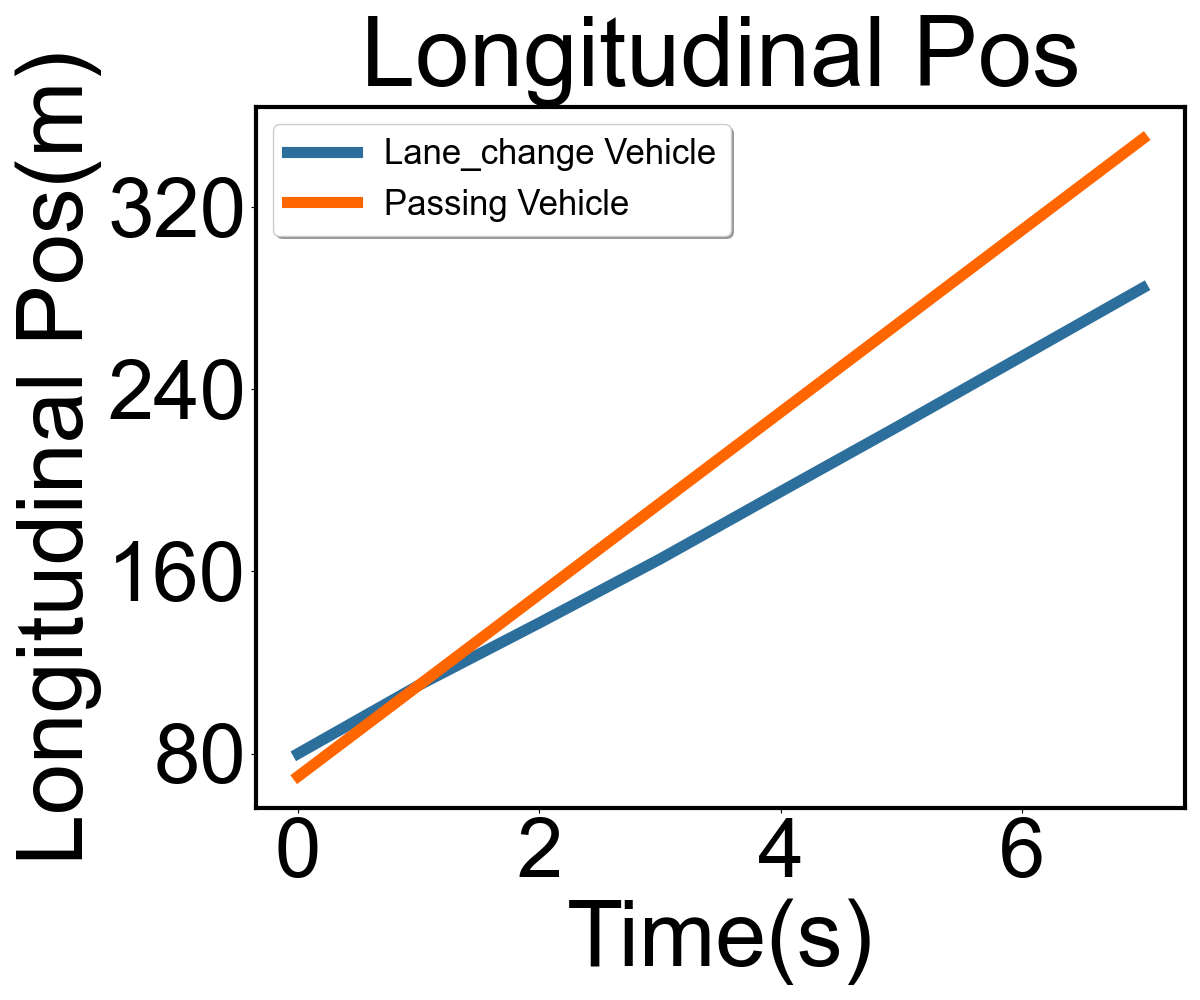} 
        \label{fig:acc}
    \end{minipage}
    \hfill
    \begin{minipage}{0.157\textwidth}
        \centering
        \includegraphics[width=\textwidth]{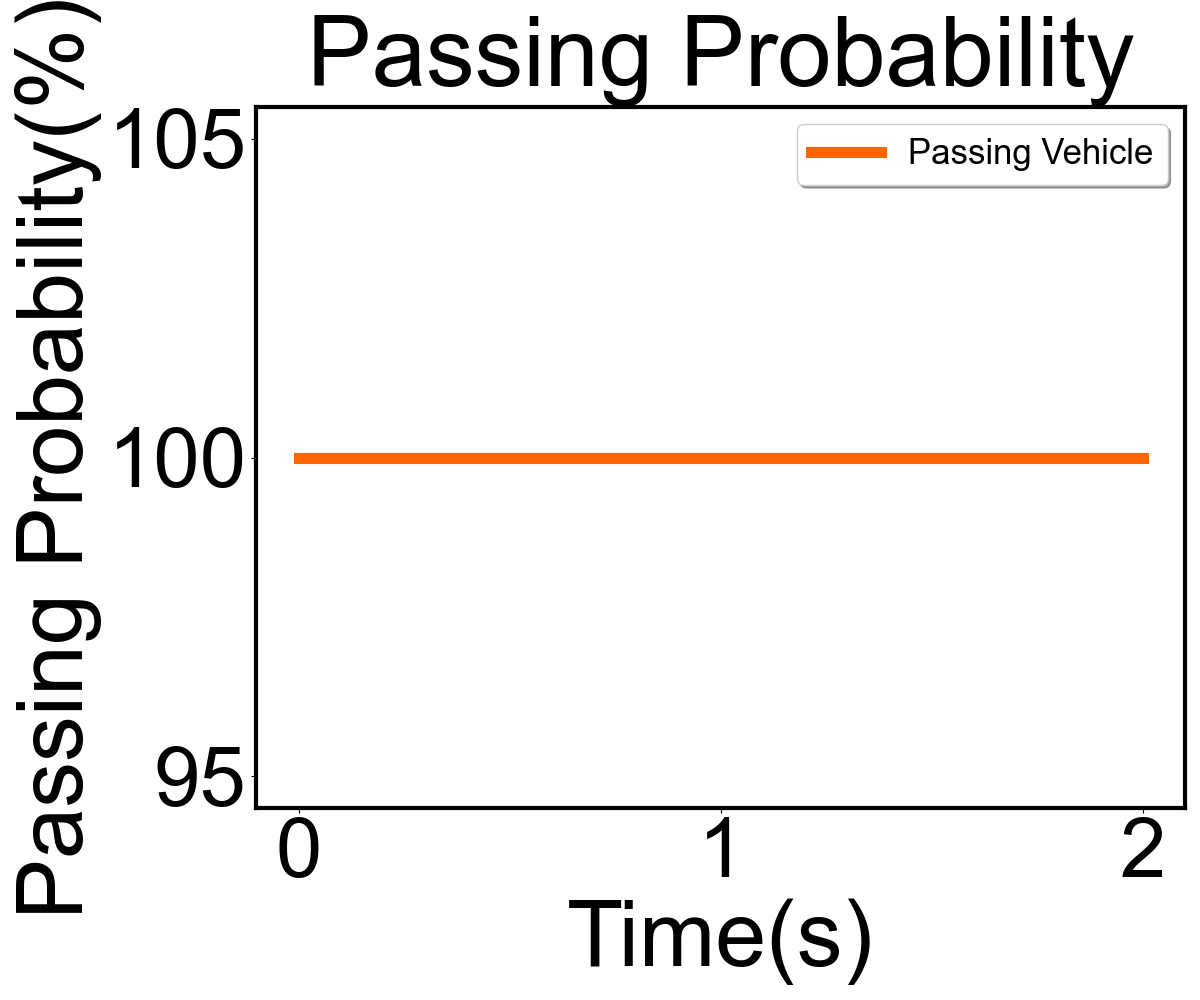} 
        \label{fig:pos}
    \end{minipage}
    \caption{Temporal Evolution of Vehicle States in Case 2.} 
    \label{fig:all_figures}
\end{figure}
The behavioral analysis reveals three key findings. First, As the AV executes its lane-changing maneuver, the passing vehicle exhibits clear passing behavior with no observable speed reduction, as demonstrated by the acceleration and velocity profiles. Second, SVO analysis confirms that the passing vehicle exhibits self-interested behavior prior to overtaking the AV. Finally, the inferred passing probability of the passing vehicle remains at 100 percent before it overtakes the AV. These results collectively confirm the surrounding vehicle's deliberate decision not to yield to the AV.

\subsubsection{Quantitative Results}
Table II provides an overview of the comparative results for the collision rate, successful rate, average reward, and average speed. 
The results demonstrate that the A2C algorithm performs the worst in terms of collision rate, successful rate and average reward. A2C can suffer from instability in learning because of the high variance in policy gradient estimation. In addition, DQN-based algorithms are typically more sample efficient than A2C. such as DQN uses an experience replay mechanism, allowing the same data sample to be used multiple times, whereas A2C is an online learning algorithm where each sample is used only once. However, the DQN algorithm and other three advanced DQN algorithms, including Double DQN, Dueling DQN and D3QN still have higher collision rate, lower average reward and successful rate, which means that these algorithms still present significant challenges in terms of safety in the dynamic environment. The social intention estimation model integrated with the DQN algorithm can significantly improves performance across key metrics. It reduces the collision rate to 7.52 percent, has the highest successful rate of 92.02 percent, achieves the highest average cumulative reward of 10.63, and maintains a high average speed of 26.20 m/s. Specifically, DQN-YI achieves the lowest collision rate, the highest cumulative reward and successful rate, and sustains a high average speed by considering the driving intentions of surrounding vehicles, which means it can balance driving efficiency and safety. 

\section{Conclusion}
In this paper, we propose a novel decision-making framework based on DRL that explicitly accounts for the driving intentions of surrounding HVs to address the challenge of high-level lane-change decision-making. Our approach comprises two key components. First, we develop a social intention estimation model by integrating SVO into a BN to represent the social preferences of surrounding vehicles. Second, we embed this estimation model within the DRL framework, enabling the decision-making process to infer and adapt to the intentions of surrounding HVs more effectively. The effectiveness of the proposed approach is validated through extensive simulations, demonstrating significant improvements in efficiency and safety. By enabling AVs to make more informed and socially-aware lane-change decisions, our framework not only improves AVs safety but also facilitates harmonious interaction with HVs.\\
Our future work will focus on developing an efficient multi-agent DRL algorithm capable of handling increasingly complex and large-scale traffic scenarios. Additionally, enhancing the robustness and adaptability of state-of-the-art decision-making algorithms is essential to ensure their effectiveness in real-world traffic environments involving multi-vehicle interactions.

\bibliographystyle{unsrt}

\bibliography{ref}

\end{document}